\documentclass[10pt, journal, letterpaper]{IEEEtran}

\usepackage{amsmath}
\usepackage{amssymb}
\usepackage{amsfonts}
\usepackage{amsthm}
\usepackage{algorithm}
\usepackage{algorithmic}
\usepackage{tikz}
\usepackage{adjustbox}
\usepackage{float}
\usepackage{verbatim}
\usepackage[caption=false]{subfig}

\begin{document}
\let\WriteBookmarks\relax
\def\floatpagepagefraction{1}
\def\textpagefraction{.001}

\title{\LARGE \bf Multi Agent Reinforcement Learning enhanced Federated Learning for Predicting Remaining Useful Lifetime}

\author{Rajarshi Chattopadhyay and Chen-Khong Tham \\Department of Electrical and Computer Engineering\\National University of Singapore\\E-mail: ceerc@nus.edu.sg, eletck@nus.edu.sg }

\maketitle

\begin{abstract}
Remaining useful life (RUL) prediction is central to Prognostics and Health Management (PHM). The growth of industrial IoT has unlocked large multivariate sensor streams and enabled deep models—especially Transformers—for long-horizon sequence learning. However, transformers are parameter-heavy and data-hungry, while operational datasets are often siloed and privacy-sensitive. Federated Learning (FL) addresses this by training across clients without raw data sharing, but in practice many clients connect over lossy wireless links where UDP is preferred for low latency, making model exchange vulnerable to packet drops.

We present a cooperative multi-agent reinforcement learning (MARL) framework that adapts FL to lossy channels by selecting per-client layer subsets to transmit and capping retransmissions. This framework is  trained with QMIX value-function factorization for scalability. We instantiate the approach with a Graph Convolution Position-Aware Transformer (GCPATr) for RUL forecasting. On the CMAPSS dataset, our MARL-enhanced FL (RL-FL) consistently improves mean square error over standard FL, with the largest gains at high packet loss (e.g., up to \textbf{18\%} improvement on FD003 at 50\% loss). These results show that MARL is an effective countermeasure to wireless degradation in FL-based RUL prediction.
\end{abstract}


\section{Introduction} \label{SecIntro}

Maintenance and prognostics play a very crucial role in modern manufacturing industry. Traditional methods such as breakdown corrective maintenance and scheduled preventive maintenance cannot satisfy the increasing demand of economic efficiency, reliability and safety. Therefore, advanced condition-based maintenance (CBM) has been proposed \cite{lee2014prognostics} to reduce maintenance cost, improving reliability and enhancing performance over long working periods. Key techniques of CBM are sensor selection for monitoring, data acquisition, data processing, algorithms for fault identification, failure prediction, remaining useful life estimation and so on. Remaining useful life (RUL) can be estimated based on historical data, which is very important for improving maintenance schedules in order to avoid engineering failures and save the resultant costs.

RUL forecasting is a multivariate time series forecasting (MVTSF) problem. MVTSF has wide applications across many domains such as intelligent transportation, predictive maintenance, weather forecasting, quantitative finance and even the spread of infection during the COVID-19 pandemic. In each of these domains, the methods used for MVTSF have evolved from simple statistical models like linear regression, Support Vector Machines (SVM) and Autoregressive Integrated Moving Average (ARIMA), to deep learning models like Recurrent Neural Networks (RNN) and its variants, and finally to transformer and its variants. Transformer \cite{vaswani2017attention} is arguably the most successful sequence modeling architecture, demonstrating unparalleled performance in various applications, such as natural language processing (NLP) \cite{devlin2018bert}, speech recognition \cite{dong2018speech} and computer vision \cite{zeng2022deciwatch}. Recently, there has also been a surge of transformer-based solutions for time series analysis, as surveyed in \cite{wen2022transformers}. Most notable models include LogTrans \cite{li2019enhancing}, Informer \cite{zhou2021informer}, Autoformer \cite{wu2021autoformer}, Pyraformer \cite{liu2021pyraformer}, Triformer \cite{cirstea2022triformer} and the recent FEDformer \cite{zhou2022fedformer}.

The essence of the transformer architecture is its multi-head self-attention mechanism. It excels in extracting semantic correlations among elements in a long sequence, e.g. words in texts/passages. Self-attention is intrinsically position or permutation invariant. Using various types of positional encoding techniques preserves some ordering information. However, there is still significant loss of ordering information especially in deep transformers where several layers of self-attention is applied. The work in \cite{chattopadhyay2024position} proposed a position aware transformer architecture (PATr) with the idea of permutation and position awareness being incorporated into the self-attention mechanism. PATr was shown to achieve state-of-the-art results in various domains. Another important observation in the case of multivariate time series data (such as those used for RUL prediction) is that it may have non-Euclidean dependencies. More precisely, we do not know about the interdependence between the multivariate time series. It is well known that the best way to model the interdependence in non-Euclidean data is to use Graph Convolution Networks (GCN) \cite{bruna2014spectral, bronstein2017geometric}. The work in \cite{chattopadhyay2025graph} propose the Graph Convolution Position Aware Transformer (GCPATr) architecture which enhances the non-Euclidean interdependency modeling power of PATr by incorporating graph convolution operations at critical places in its architecture.

A characteristic feature of all transformer architectures is that they have a large number of free parameters and require a lot of data for training. When training them for RUL forecasting, the required volume of data may not be available at one organization. Furthermore, operational data of industrial machines is private which organizations may not be willing to share. In recent years, a lot of work has been done on Federated Learning (FL). FL is a framework for communication efficient learning of deep neural networks using decentralized datasets \cite{mcmahan2017communication}. In FL, the training data resides across several client devices and model training is performed without clients explicitly sharing their private data. This framework perfectly fits the requirements our considered scenario. In addition, we consider FL clients that use UDP communication over lossy wireless networks. This is especially important for RUL forecasting of vehicular or airplane components. Their mobility restricts the usage of wired communication and the lightweight UDP protocol saves energy and reduces communication delay. The exchange of model parameters between FL clients and a centralized coordinator over lossy wireless network leads to reduction of performance of the final trained model. In this work, we propose a multi-agent reinforcement learning (MARL) based framework for mitigating the effect of packet loss. The MARL agent of each FL client intelligently selects the number of layers to be retrained during each FL round and the maximum allowed number of retransmissions when packet loss occurs. FL may involve a large number of client devices. In order to make our approach scalable, we use the idea of value function factorization to avoid action spaces with extremely large dimensions. More specifically, we use QMIX value function factorization \cite{rashid2020monotonic} which uses a more powerful and general mixing model compared to simple Value Decomposition Networks (VDN)  \cite{sunehag2017value}.

Our main contributions are summarized as follows. Firstly, we propose a cooperative MARL based framework for mitigating the effects of packet loss for FL over noisy wireless channels. Secondly, we conduct extensive simulations to validate the performance of our proposed approach in a RUL forecasting task. To the best of our knowledge, this is the first work attempting to optimize the performance of FL training of transformer-based RUL forecasting models over noisy wireless networks. The rest of the paper is organized as follows. In Section \ref{SecMethod}, we define the problem, briefly describe the GCPATr model and the cooperative MARL framework in detail. Section \ref{SecExp} describes the experimental details and presents the performance comparison of RUL forecasting models trained with and without cooperative MARL-based wireless loss mitigation. Finally, we conclude the paper in Section \ref{SecConc} and briefly discuss some future research directions.

\section{Methodology} \label{SecMethod}
\subsection{Problem Definition}
In this work, we consider a federated learning based RUL prediction setup over a noisy wireless environment. The input can be written as $X^{t} \in \mathbf{R}^{K}, \text{ } t =(1,2,..,T)$, where $T$ denotes the size of the lookback window and $K$ denotes the number of sensor readings obtained per time step. Our objective is to learn a mapping $Y_{RUL} = f(X^{t})$ for real-time RUL prediction. Here, $Y_{RUL}$ denotes the predicted RUL. We will train the GCPATr model for RUL forecasting using data distributed over many FL client devices communicating with the central server over a noisy wireless channel.

\subsection{Federated Averaging}
The Federated Averaging (FedAvg) \cite{mcmahan2017communication}, uses a central server to communicate with clients for decentralized training. Specifically, given a total of $N_{c}$ client devices, during each training round $t$, a fixed number of $N$ $(N \leq N_{c})$ devices are randomly picked by the central server. Each selected client $n$ $(1 \leq n \leq N_{c})$ contains local training data of size $D_{n}$. During the FL operation, each client device receives a copy of the global transformer model $W_{t}$ from the central server, performs $E$ epochs of local training using its local dataset, and transmits the weight updates $\Delta W_{n}^{t}$ back to the central server. The central server then computes the weighted average as shown in equation (\ref{FedAvgEq}). This process is repeated several times and each epoch is known as a communication round.

\begin{equation} \label{FedAvgEq}
    W_{glb}^{t+1} = W_{glb}^{t} + \frac{\sum_{n \in N} D_{n}.\Delta W_{n}^{t}}{\sum_{n \in N} D_{n}}
\end{equation}

\subsection{Graph Convolution Position Aware Transformer}
\begin{figure*}[!t]
\centering
\subfloat[][Overall architecture of the Graph Convolution Position Aware Transformer.]{\includegraphics[width=0.74\textwidth]{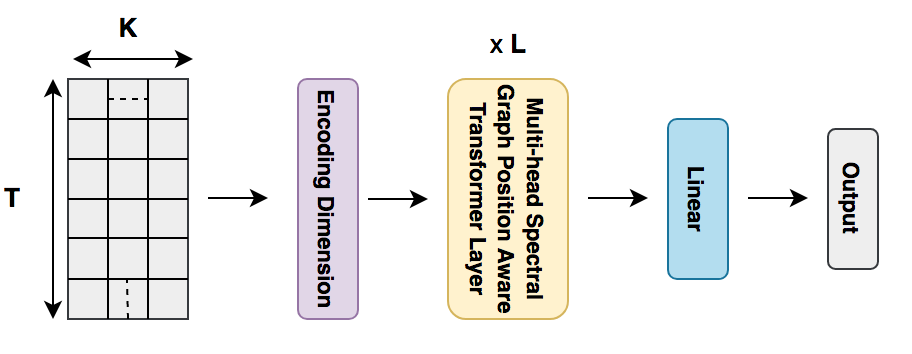}
\label{GCPATr_OA}}\\[0.45em]
\subfloat[][Multi-head Spectral Graph Position Aware Transformer layer.]{\includegraphics[width=0.90\textwidth]{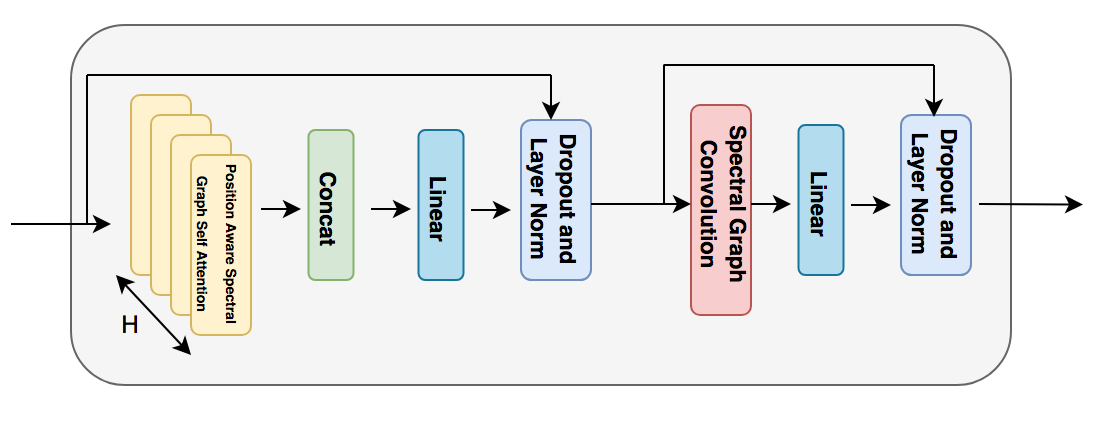}
\label{MH_GCPATr_block}}\\[0.45em]
\subfloat[][Position Aware Spectral Graph Self-Attention block.]{\includegraphics[width=0.58\textwidth]{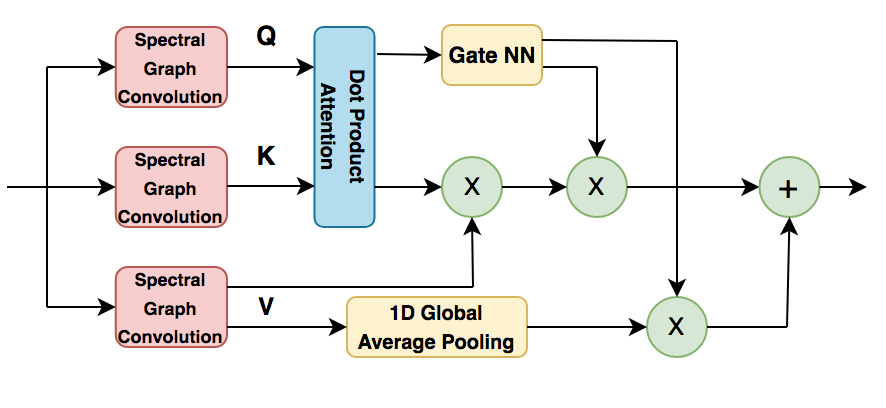}
\label{SG_PA_SA_block}}
\hfill
\subfloat[][Position Aware Self-Attention block.]{\includegraphics[width=0.36\textwidth]{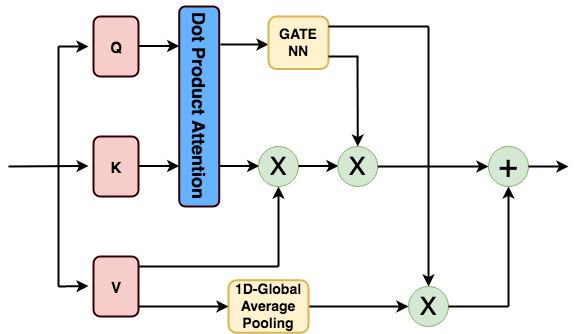}
\label{PA_SA_block}}
\caption{\label{Overall_GCPATr_Pipeline} The processing pipeline of the GCPATr model and Position Aware Self-Attention block.}
\end{figure*}

In this work, we have used the GCPATr architecture that incorporates both graph convolution and position awareness into the standard transformer architecture. The overall processing pipeline of GCPATr is shown in Figure \ref{Overall_GCPATr_Pipeline}. 

The position aware spectral graph self-attention block is shown in Figure \ref{SG_PA_SA_block} and the position aware self-attention block of PATr is shown in Figure \ref{PA_SA_block}. Different from the position aware self-attention block of PATr \cite{chattopadhyay2024position}, the linear query, key and value extractors have been replaced with spectral graph convolution operations in GCPATr. Other than this difference the working of the two blocks is the same.

The long term temporal modelling power of transformers is due to the bi-feature position invariant context modelling ability of each self-attention layer. Traditional convolution and recurrent layers can also do this, but they suffer from limited positional context and vanishing gradient issues. The self-attention mechanism overcomes these issues, but at the cost of being permutation invariant. The proposed position aware self-attention mechanism brings back the positional dependence, while still retaining the long term modelling ability of transformers. In order to achieve this, we introduce a gate neural network to the standard self-attention mechanism. This can be seen in Figure \ref{PA_SA_block}. It is a standard fully connected neural network that essentially acts as a filter and allows either the attention augmented value, or the average temporal value at that position, to pass through. The average temporal value vector contains the information summary of the entire input sequence to that layer. The filtering decision is made by the softmax activation map of the self-attention mechanism (also known as the activation pattern). The presence of a bi-feature context is not enough. The attention augmented value is only chosen if the softmax activation pattern, which has some information about the position of the two features, is of a specific type. Otherwise, the average value is returned just as it would have been if the bi-feature context was not present. This can be mathematically expressed using equations (\ref{eq_SAW}) and (\ref{eq_PASR}) below, where $\mathbf{Q}$, $\mathbf{K}$ and $\mathbf{V}$ are the query, key and value matrices, respectively. They are obtained by applying a position-wise linear transformation to the input sequence. $d_{k}$ denotes the position-wise dimensionality of the query and key.

\begin{equation} \label{eq_SAW}
    \text{W} = \text{softmax}\big(\frac{\mathbf{Q}\mathbf{K}^{T}}{\sqrt{d_{k}}}\big)
\end{equation}

\begin{equation} \label{eq_PASR}
    \text{AttentionPA}(\mathbf{Q}, \mathbf{K}, \mathbf{V}) = g(W^{T}).W.\mathbf{V} + (1-g(W^{T})).a(\mathbf{V})
\end{equation}

\begin{figure*}[!t]
        \centering \includegraphics[width=0.78\textwidth]{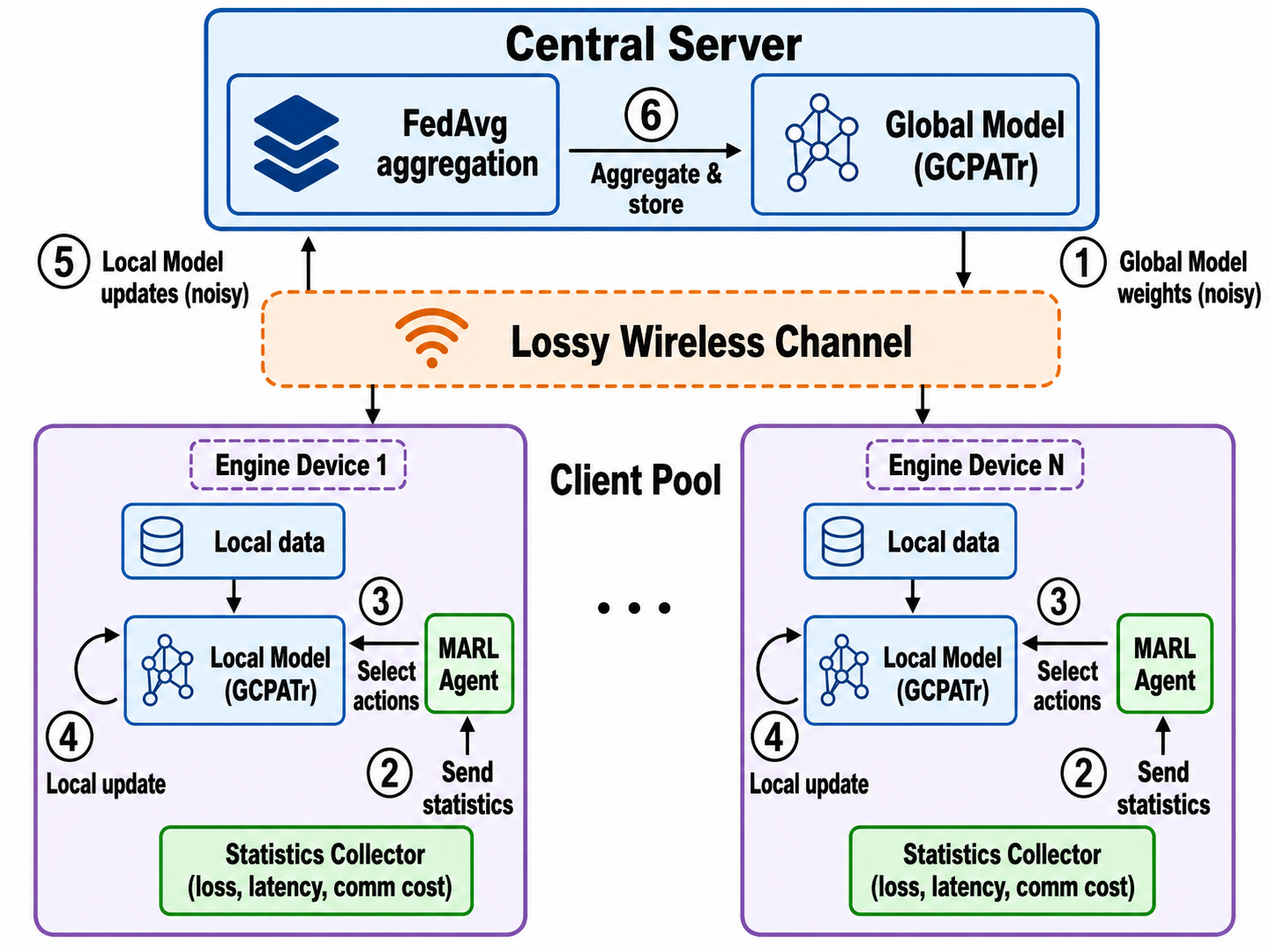}
        \caption{System diagram showing the cooperative MARL agents performing federated learning.}
        \label{sd_marl}
\end{figure*}

Here, $g(.)$ and $a(.)$ represents the gate neural network and the 1D-Global Average Pooling operation. Note that position aware self-attention mechanism is not sparse and has $O(n^{2})$ computation time and memory complexity, where $n$ denotes the length of the input sequence.


Most of the advanced deep learning models, such as Convolution Neural Networks (CNN) and Graph Convolution Networks (GCN), that specialize in modelling the interdependencies in multivariate input, require the input to have some form of structure. CNNs assume that the input features have an underlying Euclidean or regular grid-like structure. GCNs, on the other hand, are more general and its input features may have a more general graph structure. 

GCPATr does not assume that the elements of $X^{t}$ have an Euclidean graph structure. Consider the graph $G=(V, E, A)$. Here, $V$ denotes the set of nodes. Each node denotes a sensor. Since there are $K$ sensors, we have $|V| = K$. The set of edges in $G$ is denoted by $E$. An edge will exist between two nodes if there is some dependency between the respective input time series. Here, $A$ denotes the adjacency matrix. It is a $K \times K$ square matrix, with $a_{ij}$ (the $i^{\rm{th}}$ row and $j^{\rm{th}}$ column element) denoting the amount of dependency between the respective input time series. An effective way of estimating the amount of dependency between two time series is by computing their covariance. We can compute the respective covariances using the training data. Our choice of adjacency matrix is shown in equation (\ref{CovMatCalc}) below. Here, $X_{i} = [X^{1}_{i}, X^{2}_{i},...., X^{T}_{i}]$, $E[.]$ denotes the expectation operator, and ${\rm{Tr}}(X_{j})$ denotes the transpose of $X_{j}$. The expectation is taken over the entire training data. Intuitively, $a_{ij} \neq 0$ means that the reading from sensors denoted by nodes $i$ and $j$ are not independent. This becomes clear if we consider each sensor output as a random variable.


\begin{equation} \label{CovMatCalc}
    a_{ij} = E[X_{i}.(X_{j})^{T}]
\end{equation}

\subsection{Noisy Wireless Communication}
In the scenario considered, the FL client devices communicate with the central server over a noisy wireless communication channel. This implies that occasionally, parts of the model weights will be lost when transmitting the global model to client devices or when receiving model updates from client devices. We model this loss in terms of a packet drop rate $P_{dr}$. More specifically, $P_{dr}$ denotes the probability that a particular weight of the model is dropped during transmission. A dropped weight is handled in the following way:

\begin{enumerate}
    \item If a weight is dropped when broadcasting from the central server to a client device, we reuse its previous value at the client.
    \item If a weight is dropped when transmitting it from a client to the central server, we do not include it in the weighted average. The denominator term corresponding to the client whose update is lost is also ignored in the summation.
\end{enumerate}

\subsection{Cooperative Multi-Agent Reinforcement Learning (MARL)}

In cooperative MARL, a set of $N$ agents are trained to produce the optimal actions that lead to the maximum team reward. Specifically, at each timestamp $t$, each agent $n \text{ } (1 \leq n \leq N)$ observes its state $s^{t}_{n}$ and selects an action $a^{t}_{n}$ based on $s^{t}_{n}$. After all the agents complete their actions, the team receives a joint reward $r^{t}$ and proceeds to the next state $s^{t+1}$. The goal is to maximize the total expected discounted reward $R =  \sum_{t=1}^{T} \gamma^{t}.r^{t}$ by selecting the optimal agent actions, where $\gamma \in [0, 1]$ is the discount factor.

In deep cooperative MARL, a separate value function approximator $Q_{n}(s^{t}_{n}, a^{t-1}_{n})$ is learned for each participating agent. We apply the QMIX value function factorization approach \cite{rashid2020monotonic} to obtain the joint state-action value function by mixing the output of the individual client action value functions. In QMIX, a hypernetwork is used to mix the outputs of $Q_{n}(s^{t}_{n}, a^{t-1}_{n})$. The hypernetwork parameters are generated by another network which takes the global state $S^{t}$ as input. The hypernetwork parameters are generated in such a way that the mixed output is strictly monotonic w.r.t. the inputs. We will use $HN(.)$ to denote the hypernetwork. Therefore, the joint action-value can be denoted as $Q_{tot}(S^{t}, A^{t}) = HN(Q_{1}(s^{t}_{1}, a^{t-1}_{1}),...Q_{N}(s^{t}_{N}, a^{t-1}_{N}); S^{t})$.

\subsection{MARL problem formulation}
We use cooperative MARL to mitigate the effects of packet loss on the performance of our federated RUL prediction model training. Figure \ref{sd_marl} depicts the setup. More specifically, we apply MARL in the following way:

\begin{enumerate}
    \item We train one MARL agent corresponding to each FL client. We will refer to them as client agents. Each client agent picks two types of actions.
    \item Firstly, each client agent decides how many maximum retransmissions $R^{t}_{n}$ are allowed for each weight corresponding to this client, where $R^{t}_{n} \in [0, 1, 2, 3]$. Retransmission only happens in the case of packet drop and the maximum number of retransmissions are capped at $3$. Note that $R^{t}_{n} = 0$ means that client FL $n$ does not participate in round $t$.
    \item Secondly, each client agent selects the fraction of layers of the local model $ls^{t}_{n}$ to be updated in this communication round, where $ls^{t}_{n} \in [1, 2, 3, 4]$. Here, $1$ represents the bottom quarter of the layers, $2$ represents the bottom half and so on.
\end{enumerate}

Allowing retransmissions and restricting the volume of information sent over noisy communication channel gives the FL clients some power to mitigate the effect of packet loss.

The client agent observation consists of the following:
\begin{enumerate}
    \item $L^{t}_{n}$ : Probing loss of client $n$ at time step $t$, which is the value of the training loss function using the local model's current weights before updating with global model weights.
    \item $H^{u}_{t-1,n}$ : Communication latency at the previous time-step for uploading the client model to the central server.
    \item $H^{r}_{t-1,n}$ : Total number of UDP packets used by agent $n$ to upload the local model to the central server in the previous time step.
    \item $H^{l}_{t-1,n}$ : Total number of layers selected for agent $n$ at the previous time step.
    \item $D_{n}$ : Data size at client $n$.
\end{enumerate}

$S_{n}^{t} = [L^{t}_{n}, H^{u}_{t-1,n}, H^{r}_{t-1,n}, H^{l}_{t-1,n}, D_{n}]$ denotes the state observation of client agent $n$. The global state observation is simply the concatenation of all the client agent state observations. 

The reward function used in all experiments is given by equation (\ref{groupReward}), which balances accuracy improvement with communication efficiency. Here $ac(R^{t}_{i})$ denotes the actual number of retransmissions used by client $i$ at round $t$.

\begin{equation}  \label{groupReward}
\begin{aligned}
    r^{t} = {} & \frac{1}{W_{a}}\left[\sum_{i=t-W_{a}}^{t}\big(MSE_{val}(t-1) - MSE_{val}(t)\big)\right] \\
    & {} - W_{c}\sum_{i=1}^{N} ac(R^{t}_{i})\,ls^{t}_{i}
\end{aligned}
\end{equation}


Algorithm \ref{MFed_Avg_alg} shows one step of the modified FedAvg algorithm incorporating the decisions made by client agents. Here, $GM(.)$ and $LM_{i}(.)$ denote the FL global and client models corresponding to FL client node $i$, respectively. Note that these models are GCPATr transformer models in our case. Algorithm \ref{QMIX_train_alg} shows the steps involved in training a QMIX value function factorization-based MARL framework. Here, $HN(.)$ denotes the hypernetwork as described earlier.

\begin{algorithm}
\caption{Modified Step of FedAvg Algorithm $FA(R^{t}, ls^{t}, E, B)$}
\label{MFed_Avg_alg}
\begin{algorithmic}[1]

\FOR{$i=1,2,..,N$}
\IF{$R^{t}_{i} > 0$}
\FOR{$k=1,2,..,ls^{t}_{i}$}
\STATE Copy layer $k$ from $GM(.)$.
\ENDFOR
\STATE Locally train $LM_{i}(.)$ for $E$ epochs with batch size $B$.
\FOR{$k=1,2,..,ls^{t}_{i}$}
\STATE Transfer layer $k$ from $LM_{i}(.)$ to $GM(.)$, allowing up to $R^{t}_{i}$ re-transmissions.
\ENDFOR
\ENDIF
\ENDFOR
\STATE Average the received layer weights to produce updated $GM(.)$
\end{algorithmic}
\end{algorithm}

\begin{algorithm} 
\caption{QMIX MARL Training Algorithm}
\label{QMIX_train_alg}
\begin{algorithmic}[2]

\STATE Initialize $GM(.)$ randomly and copy to $LM_{i}(.) \forall i \in (1,2,...,N)$.
\STATE Initialize $Q_{i}(.); \forall i \in (1,2,...,N)$ randomly.

\FOR{$t = 1,2,...$}
\STATE Compute $L^{t}_{n}$ and prepare $S^{t}_{i}; \forall i \in (1,2,...,N)$ .
\STATE Determine actions using epsilon-greedy policy, i.e. $R^{t}_{i}, ls^{t}_{i} = \pi_{\epsilon}(Q^{i}(S^{t}_{i}, a^{t-1}_{i}))$
\STATE Do $FA(R^{t}, ls^{t}, E, B)$.
\STATE Compute Reward function $r^{t}$ and $S^{t+1}_{i}$ for each agent.
\STATE Store the experience $<S^{t}, a^{t}, r^{t}, S^{t+1}>$ into replay memory.

\STATE Compute $Q_{tot}(S^{t}, A^{t}) = HN(Q_{1}(s^{t}_{1}, R^{t}_{1}, ls^{t}_{1}),..., Q_{N}(s^{t}_{N}, R^{t}_{N}, ls^{t}_{N}); S^{t})$ 

\STATE Compute loss function $Loss(t) = \sum_{z=1}^{b}(r + \gamma.\max_{a}Q^{tot}(s^{t+1}, a, \theta^{-}) - Q^{tot}(s^{t}, a^{t}, \theta))$, using $b$ samples from replay memory.
\STATE Update $\theta$ using backpropagation on $Loss(t)$.
\STATE Set $\theta^{-} = \theta$ once every $T_{tgn}$ time steps.
\ENDFOR
\end{algorithmic}
\end{algorithm}

\section{Experimental Details and Parameter Setup} \label{SecExp}
In this section, we describe our experimental setup. We have used the CMAPSS dataset \cite{saxena2008damage} to do RUL forecasting. The CMAPSS dataset consists of simulated flight turbofan data from $100$ turbofans operating under different conditions. To emulate a non IID dataset, each FL client dataset contains a small amount of data from each of the $100$ turbofans and with the majority of the data coming from a subset of turbofans, i.e. a unique subset chosen for each client with no overlap. We emulate the noisy communication channel by considering different packet loss probabilities between $0$ and $0.5$. We set the total number of available FL clients to $N=20$, number of epochs per communication round $E=3$ and batch size $B=64$. The number of client agents is the same as the number of FL clients. The RL agents were trained for $300$ episodes with each episode running the FL training for $500$ communication rounds. The QMIX mixture model and client agents were trained using a replay memory with size $500$ and batch size $64$. $W_{c}$ was set to $0.01$ and while training, a context length of $5$ was used for the client agent GRU. Lastly, $\lambda = 0.95$ and the epsilon greedy parameter was set to $0.9$ at the start and gradually decreased to $0.1$. The performance of the proposed RUL prediction model was measured using the moving average of Mean Square Error (MSE) metric.

\section{Results and Performance Analysis} \label{SecResults}

\noindent\textbf{FD001.}
Table \ref{tab:fd001} presents the performance comparison under different packet loss rates. At $0\%$ loss, RL-FL slightly improves over FL (18.65 vs.\ 18.81). At $10\%$ loss, the two are comparable (18.58 vs.\ 18.07). At higher loss rates, the benefits become clear: at $30\%$ loss, RL-FL reduces MSE from 20.52 to 19.85 (\textbf{+3.3\%}), and at $50\%$ loss, from 22.36 to 19.73 (\textbf{+11.8\%}). This indicates that MARL is especially effective under severe packet loss.

\noindent\textbf{FD002.}
RL-FL consistently improves across all packet loss levels. At $0\%$ loss, MSE drops from 32.12 to 30.51 (\textbf{+5.0\%}). At $10\%$, the improvement is \textbf{+7.1\%}; at $30\%$, \textbf{+6.7\%}; and at $50\%$, \textbf{+11.1\%}. FD002 shows robust gains for RL-FL, especially at higher loss.

\noindent\textbf{FD003.}
RL-FL provides major gains at moderate and high losses. At $0\%$ loss, it is comparable (24.38 vs.\ 24.17). At $10\%$, RL-FL reduces MSE from 24.88 to 22.22 (\textbf{+10.7\%}). At $30\%$, the improvement is \textbf{+12.5\%}, and at $50\%$, \textbf{+18.0\%}, the strongest across datasets.

\noindent\textbf{FD004.}
Results are mixed. At $0\%$ loss, RL-FL is worse (47.25 vs.\ 43.71). At $10\%$, it improves performance (\textbf{+5.8\%}), while at $30\%$ it is slightly worse (46.80 vs.\ 44.89). At $50\%$, RL-FL again improves (43.78 vs.\ 44.77, \textbf{+2.2\%}). This suggests dataset-specific sensitivity, but still shows advantages under high packet loss.

\begin{table}[h!]
\caption{Performance (MSE) achieved by FL and MARL-enhanced FL (RL-FL) on FD001.}
\label{tab:fd001}
\centering
\begin{tabular}{|c||c|c|c|}
\hline
Channel Loss & FL & Random Action & RL-FL \\
\hline\hline
$0\%$  & 18.81 & \textbf{18.64} & 18.65 \\
$10\%$ & \textbf{18.07} & 19.21 & 18.58 \\
$30\%$ & 20.52 & \textbf{19.35} & 19.85 \\
$50\%$ & 22.36 & 21.03 & \textbf{19.73} \\
\hline
\end{tabular}
\end{table}

\begin{table}[h!]
\caption{Performance (MSE) achieved by FL and MARL-enhanced FL (RL-FL) on FD002.}
\label{tab:fd002}
\centering
\begin{tabular}{|c||c|c|c|}
\hline
Channel Loss & FL & Random Action & RL-FL \\
\hline\hline
$0\%$  & 32.12 & 31.84 & \textbf{30.51} \\
$10\%$ & 32.61 & 32.41 & \textbf{30.31} \\
$30\%$ & 34.35 & 32.52 & \textbf{32.06} \\
$50\%$ & 33.97 & 32.65 & \textbf{30.20} \\
\hline
\end{tabular}
\end{table}

\begin{table}[h!]
\caption{Performance (MSE) achieved by FL and MARL-enhanced FL (RL-FL) on FD003.}
\label{tab:fd003}
\centering
\begin{tabular}{|c||c|c|c|}
\hline
Channel Loss & FL & Random Action & RL-FL \\
\hline\hline
$0\%$  & \textbf{24.17} & 24.47 & 24.38 \\
$10\%$ & 24.88 & 22.34 & \textbf{22.22} \\
$30\%$ & 25.37 & 24.07 & \textbf{22.19} \\
$50\%$ & 26.90 & 25.43 & \textbf{22.06} \\
\hline
\end{tabular}
\end{table}

\begin{table}[h!]
\caption{Performance (MSE) achieved by FL and MARL-enhanced FL (RL-FL) on FD004.}
\label{tab:fd004}
\centering
\begin{tabular}{|c||c|c|c|}
\hline
Channel Loss & FL & Random Action & RL-FL \\
\hline\hline
$0\%$  & \textbf{43.71} & 44.68 & 47.25 \\
$10\%$ & 46.12 & 45.12 & \textbf{43.46} \\
$30\%$ & 44.89 & \textbf{44.24} & 46.80 \\
$50\%$ & 44.77 & 45.92 & \textbf{43.78} \\
\hline
\end{tabular}
\end{table}

\begin{figure*}[!ht]
\centering

\subfloat[FD001 -- Standard FL.]{
    \includegraphics[width=0.23\textwidth]{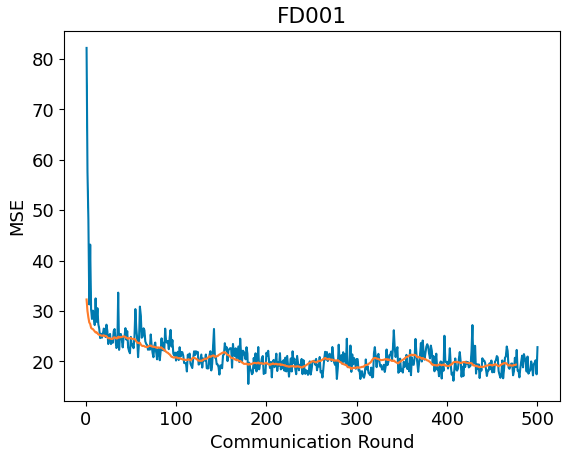}
    \label{fig:fd001_fl}
}
\hfill
\subfloat[FD001 -- RL-FL.]{
    \includegraphics[width=0.23\textwidth]{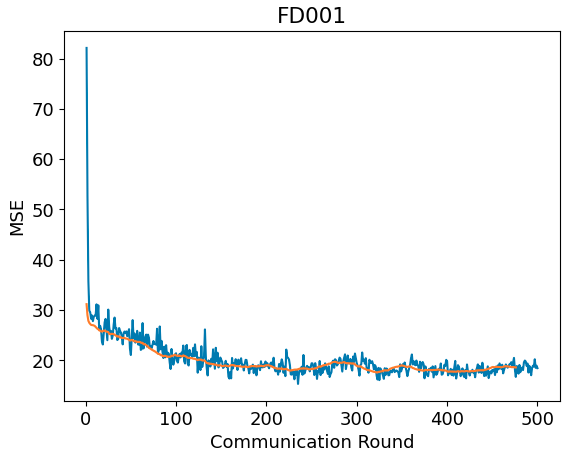}
    \label{fig:fd001_rlfl}
}
\hfill
\subfloat[FD002 -- Standard FL.]{
    \includegraphics[width=0.23\textwidth]{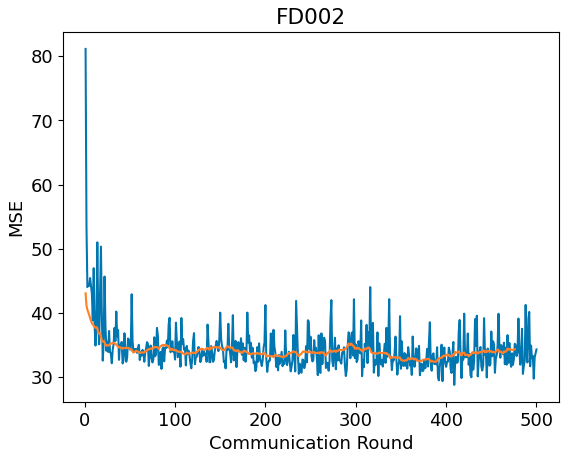}
    \label{fig:fd002_fl}
}
\hfill
\subfloat[FD002 -- RL-FL.]{
    \includegraphics[width=0.23\textwidth]{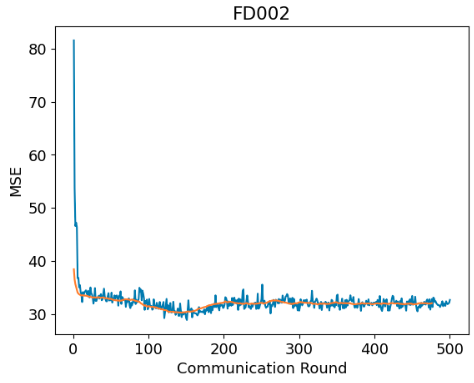}
    \label{fig:fd002_rlfl}
}

\vspace{0.5em}

\subfloat[FD003 -- Standard FL.]{
    \includegraphics[width=0.23\textwidth]{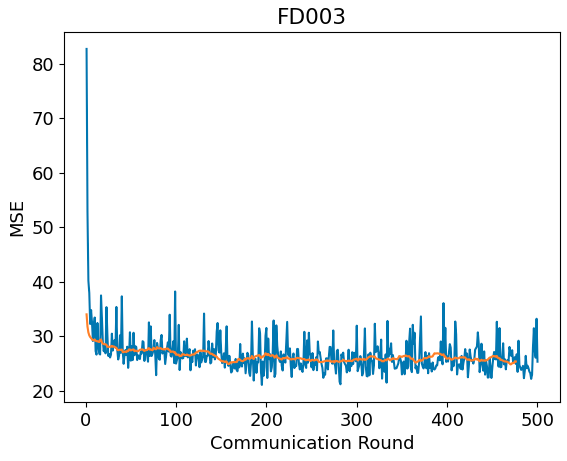}
    \label{fig:fd003_fl}
}
\hfill
\subfloat[FD003 -- RL-FL.]{
    \includegraphics[width=0.23\textwidth]{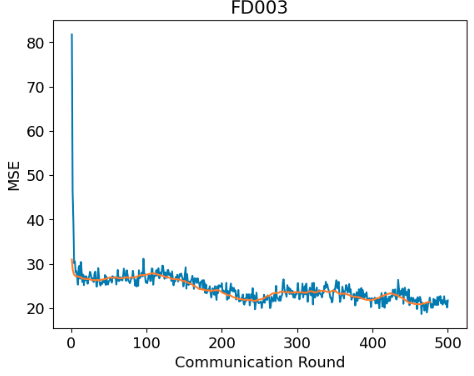}
    \label{fig:fd003_rlfl}
}
\hfill
\subfloat[FD004 -- Standard FL.]{
    \includegraphics[width=0.23\textwidth]{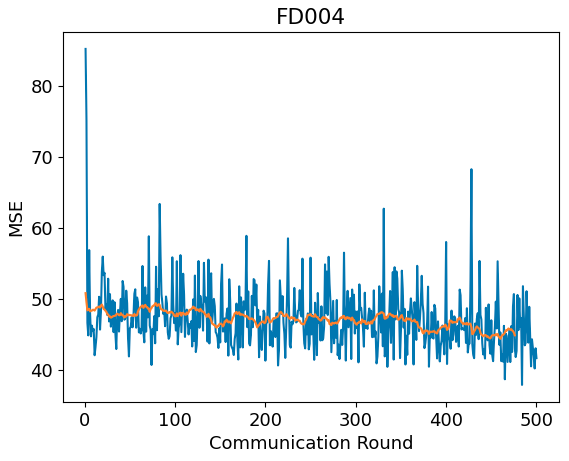}
    \label{fig:fd004_fl}
}
\hfill
\subfloat[FD004 -- RL-FL.]{
    \includegraphics[width=0.23\textwidth]{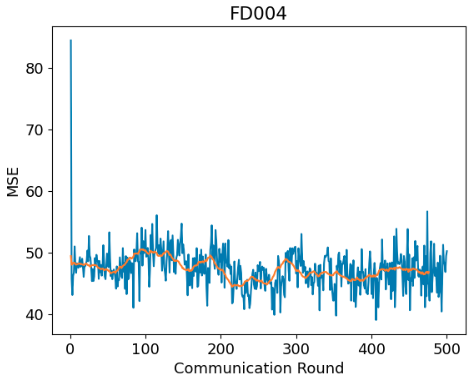}
    \label{fig:fd004_rlfl}
}

\caption{Convergence plots of Standard FL and RL-FL (MARL-enhanced FL) on \textbf{FD001--FD004} under a $30\%$ packet loss rate.}
\label{fig:all_fd_30}
\end{figure*}

In addition to comparing overall model performance across datasets, we also analyze the actions selected by the RL-FL client agents. Specifically, we study retransmission ($R^{t}_{n}$) and layer selection ($ls^{t}_{n}$) behaviors during training. We first present results on FD001 alone (Figures \ref{Plots_Rtx} and \ref{Plots_ls}), and then extend the comparison across all four CMAPSS subsets at a fixed 30\% packet drop rate (Figures \ref{Plots_rtx_across_datasets_30} and \ref{Plots_ls_across_datasets_30}).

\subsection{Analysis of Retransmission Actions}
\begin{figure*}[!t] 
\centering
\subfloat[][$10\%$ packet loss rate.]{\includegraphics[width=0.31\textwidth]{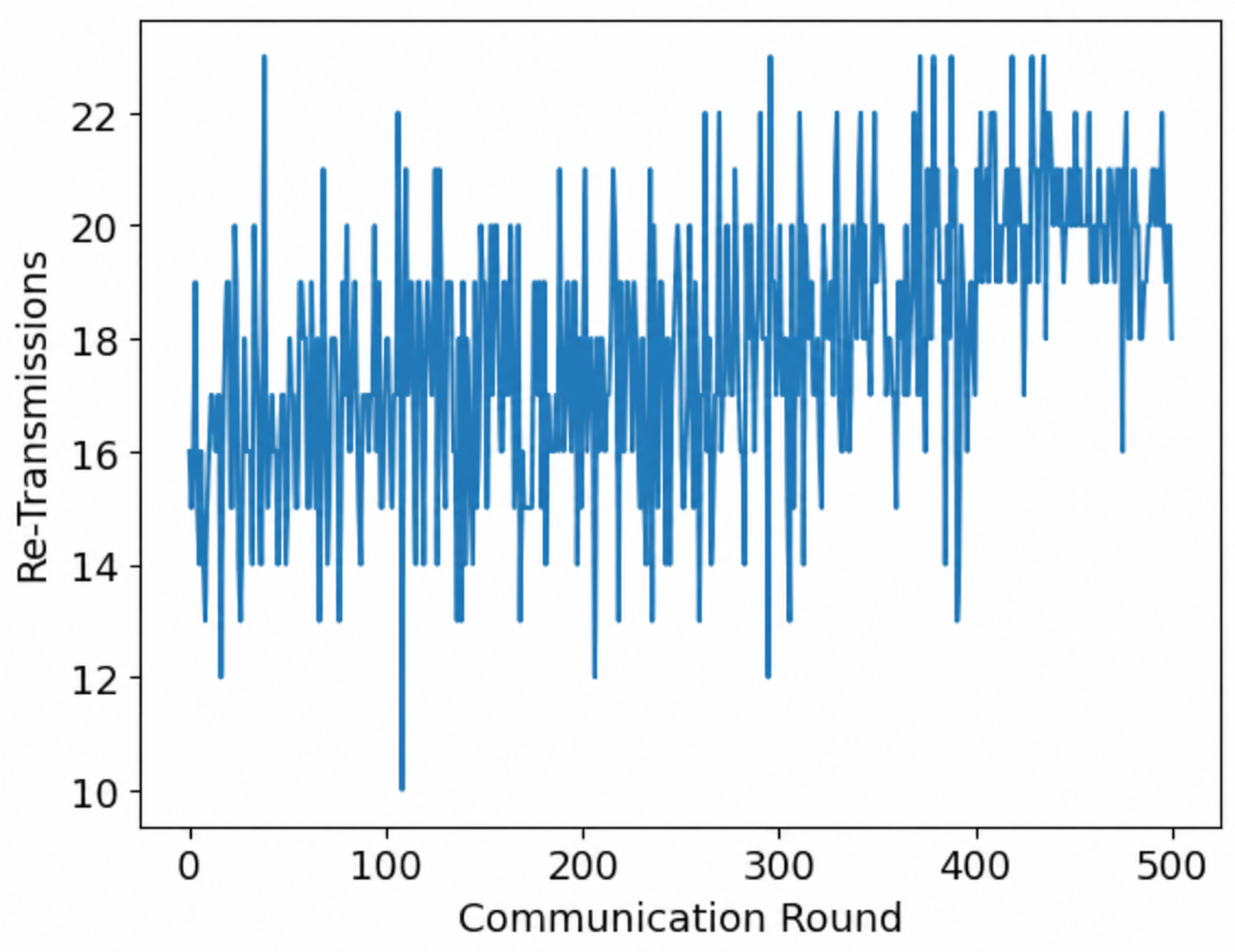}
\label{rtx10}}
\hfill
\subfloat[][$30\%$ packet loss rate.]{\includegraphics[width=0.31\textwidth]{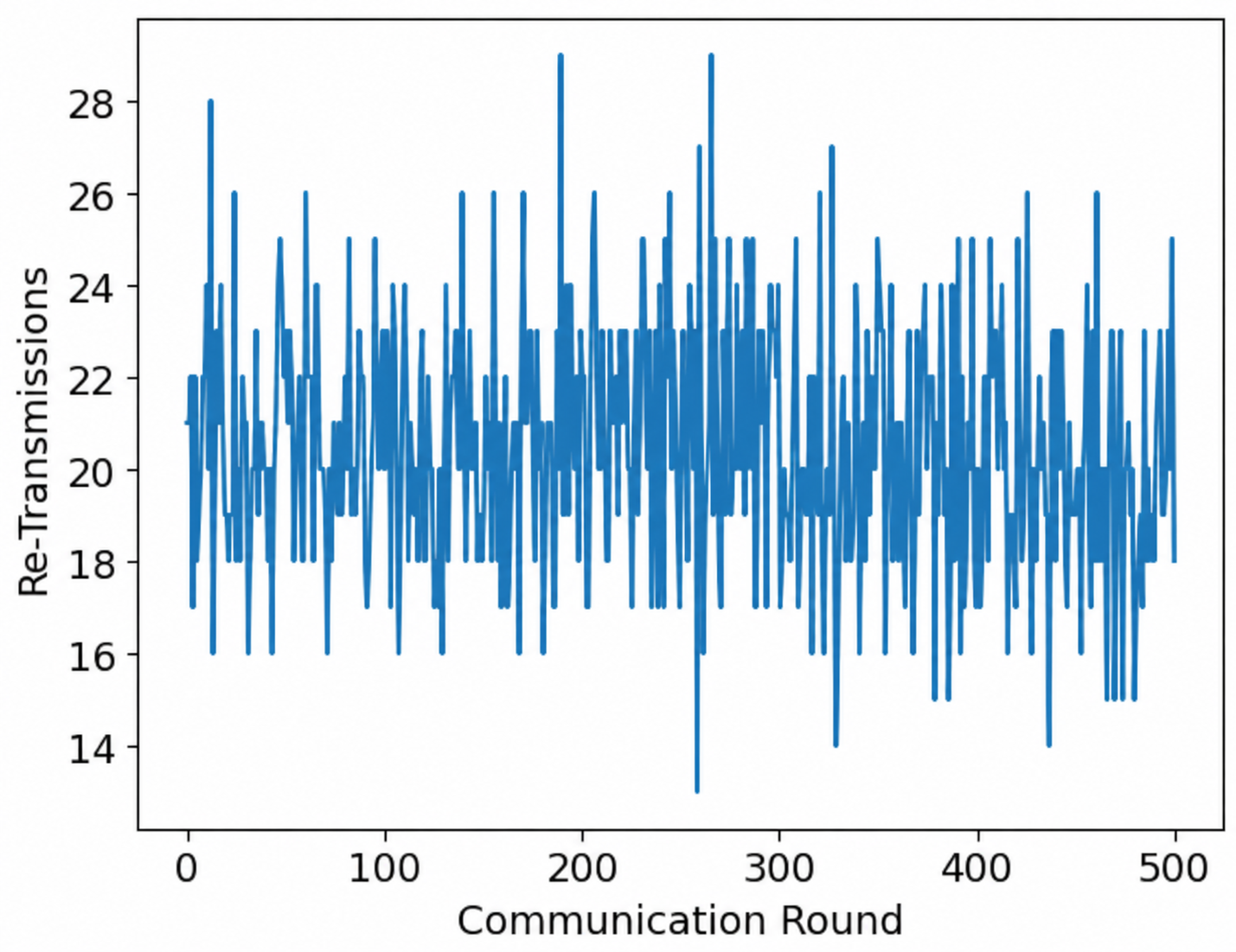}
\label{rtx30}}
\hfill
\subfloat[][$50\%$ packet loss rate.]{\includegraphics[width=0.31\textwidth]{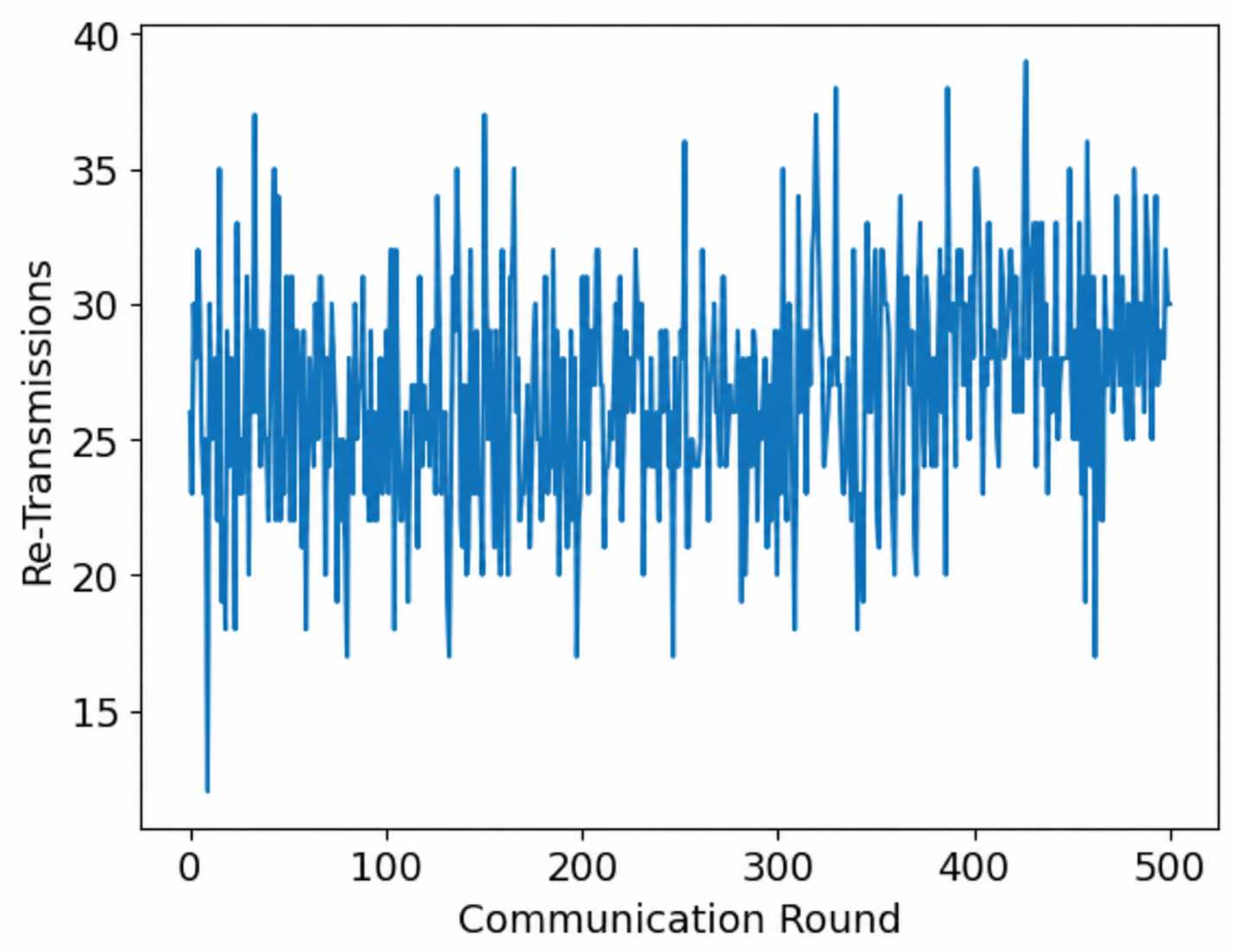}
\label{rtx50}}
\hfill
\caption{\label{Plots_Rtx} Retransmission actions selected by RL-FL client agents during FL training on \textbf{FD001} for $10\%$, $30\%$, and $50\%$ packet loss rates.}
\end{figure*}

Fig.~\ref{Plots_Rtx} shows retransmission actions on FD001 at 10\%, 30\%, and 50\% packet loss. Two clear trends emerge. First, retransmission counts increase at later communication rounds, reflecting the RL agents’ strategy to allocate more retransmissions as convergence slows. Second, higher loss rates lead to higher overall retransmission counts: the curves at 10\%, 30\%, and 50\% loss form an increasing hierarchy.

\begin{figure*}[!t]
\centering
\subfloat[][FD001]{
    \includegraphics[width=0.23\textwidth]{Rtx_30_N_N.png} 
    \label{fd001_rtx_30}
}
\hfill
\subfloat[][FD002]{
    \includegraphics[width=0.23\textwidth]{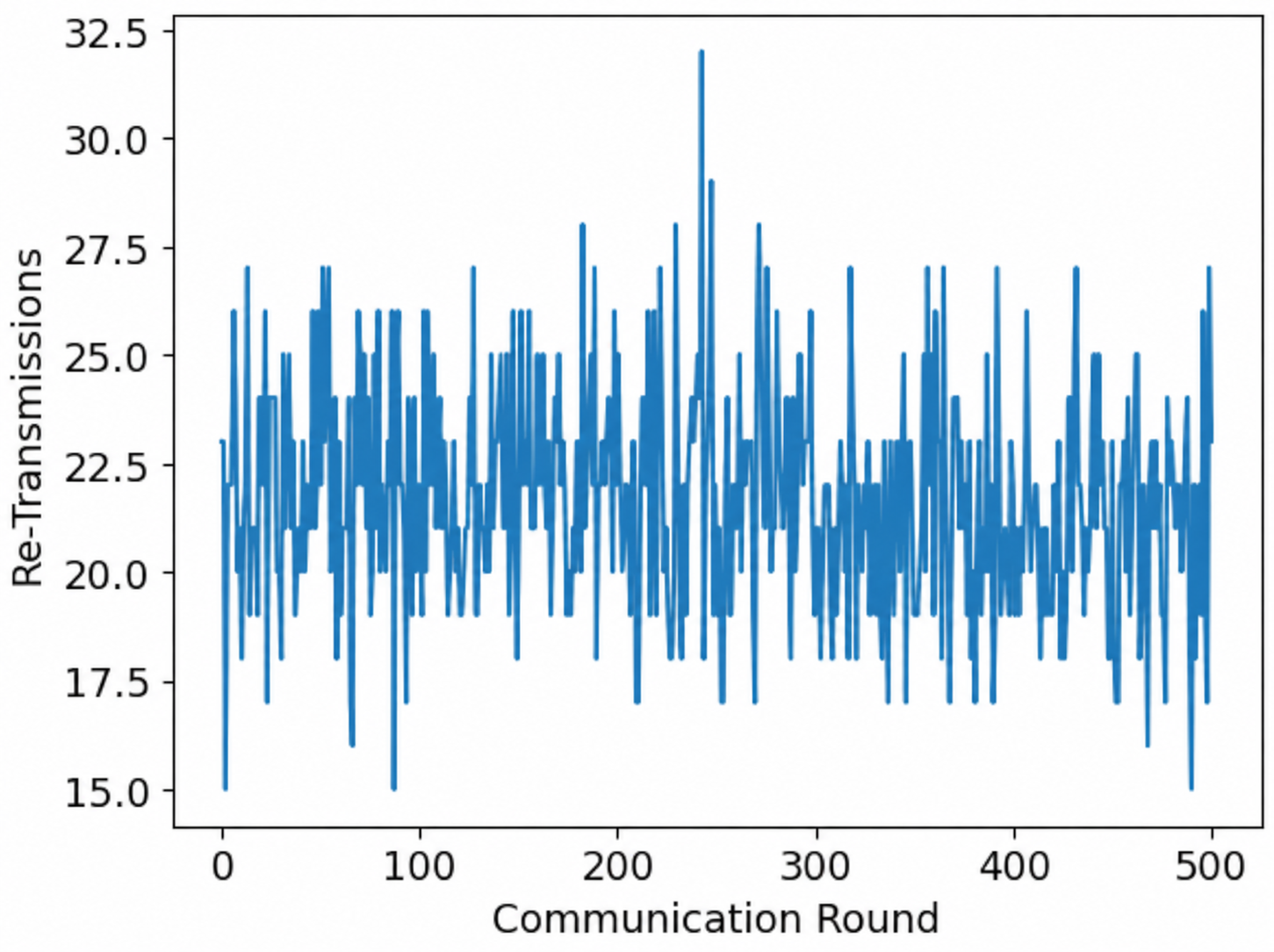} 
    \label{fd002_rtx_30}
}
\hfill
\subfloat[][FD003]{
    \includegraphics[width=0.23\textwidth]{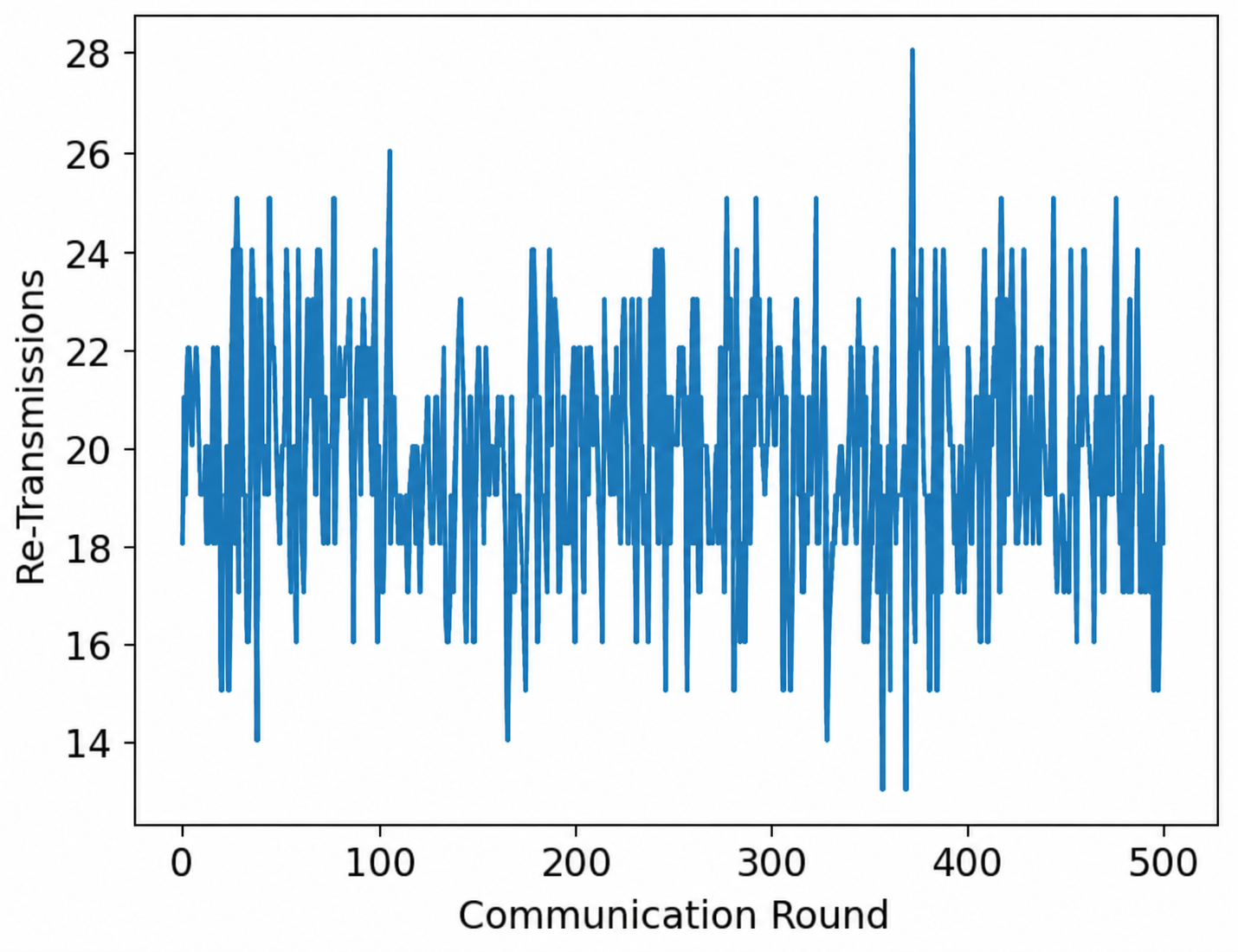} 
    \label{fd003_rtx_30}
}
\hfill
\subfloat[][FD004]{
    \includegraphics[width=0.23\textwidth]{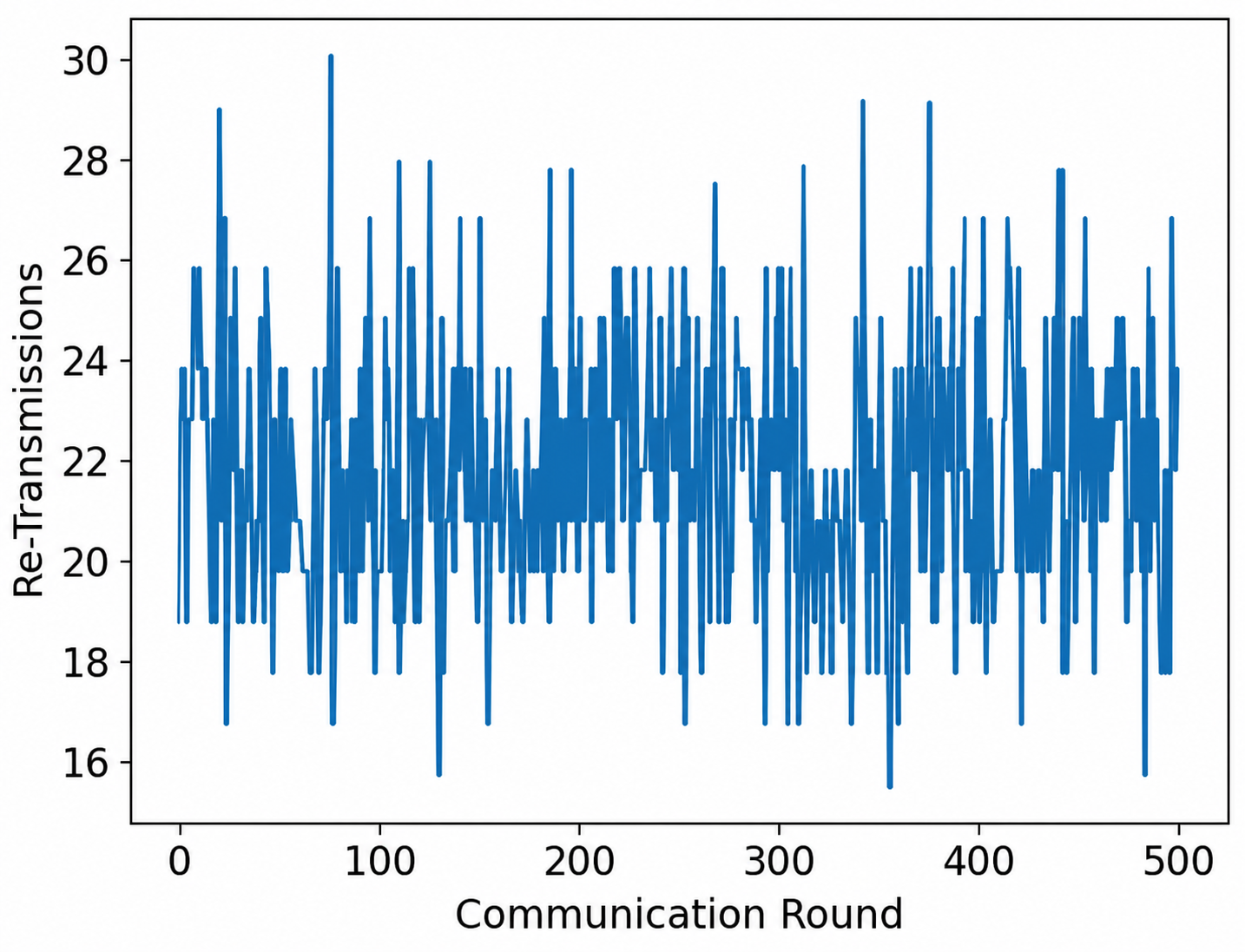} 
    \label{fd004_rtx_30}
}
\caption{\label{Plots_rtx_across_datasets_30}
Comparison of retransmission actions selected by RL-FL client agents across \textbf{FD001–FD004} at a packet loss rate of \textbf{30\%}. Each subplot shows the evolution of $\sum_{n=1}^{20} R_n^t$ over communication rounds for the corresponding dataset.}
\end{figure*}

Fig.~\ref{Plots_rtx_across_datasets_30} extends this analysis across FD001--FD004 at a fixed 30\% packet loss rate. FD001 and FD003 show steadily increasing retransmissions, but their overall values are visibly lower compared to FD002. FD002 maintains a flatter trajectory but at a consistently higher level, suggesting that more retransmissions are needed to stabilize learning. FD004 exhibits considerable noise, making cross-dataset comparison less conclusive. These results highlight that dataset-specific factors such as sensor redundancy and covariance structure strongly influence how retransmission policies are learned.

\subsection{Analysis of Layer Selection Actions}
\begin{figure*}[!t] 
\centering
\subfloat[][$10\%$ packet loss rate.]{\includegraphics[width=0.31\textwidth]{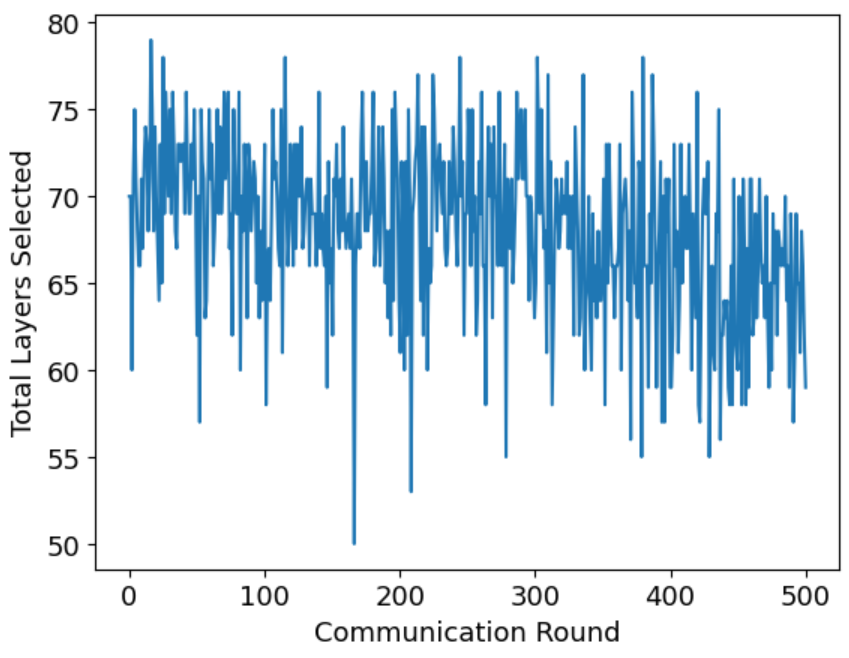}
\label{ls10}}
\hfill
\subfloat[][$30\%$ packet loss rate.]{\includegraphics[width=0.31\textwidth]{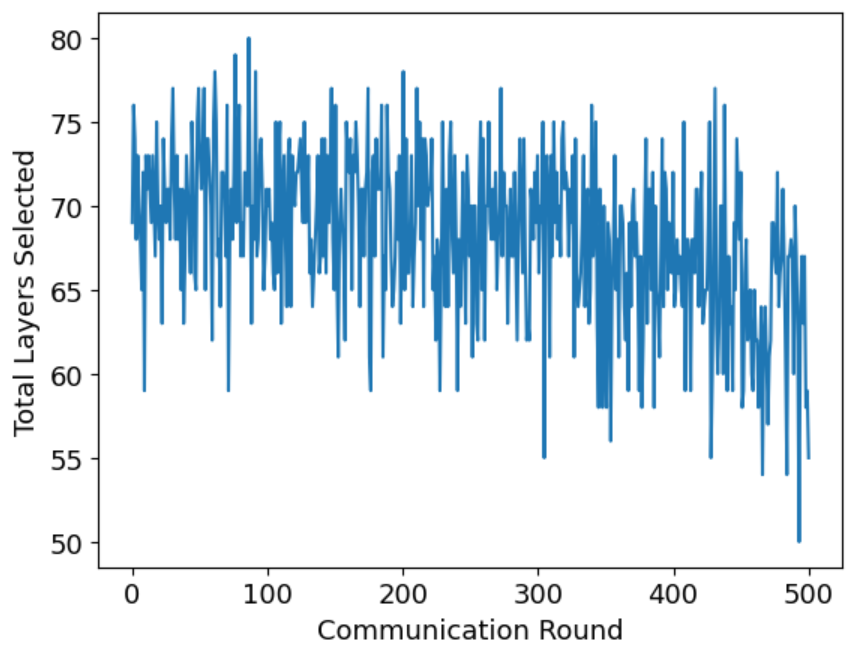}
\label{ls30}}
\hfill
\subfloat[][$50\%$ packet loss rate.]{\includegraphics[width=0.31\textwidth]{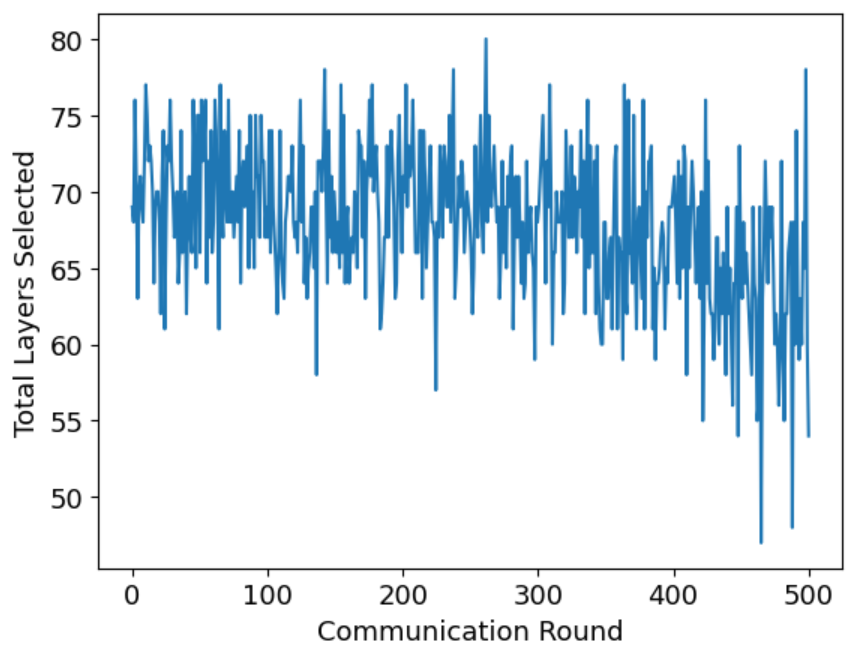}
\label{ls50}}
\hfill
\caption{\label{Plots_ls} Layer selection actions chosen by RL-FL client agents during FL training on \textbf{FD001} for $10\%$, $30\%$, and $50\%$ packet loss rates.}
\end{figure*}

Fig.~\ref{Plots_ls} shows layer selection actions on FD001 at 10\%, 30\%, and 50\% packet loss. Unlike retransmissions, the total number of layers selected remains relatively stable across loss rates, with only a slight decline at later rounds. This reflects the fact that once the models are close to convergence, updating fewer layers is sufficient.

\begin{figure*}[!t]
\centering
\subfloat[][FD001]{
    \includegraphics[width=0.23\textwidth]{Ls_30_N_N.png} 
    \label{fd001_ls_30}
}
\hfill
\subfloat[][FD002]{
    \includegraphics[width=0.23\textwidth]{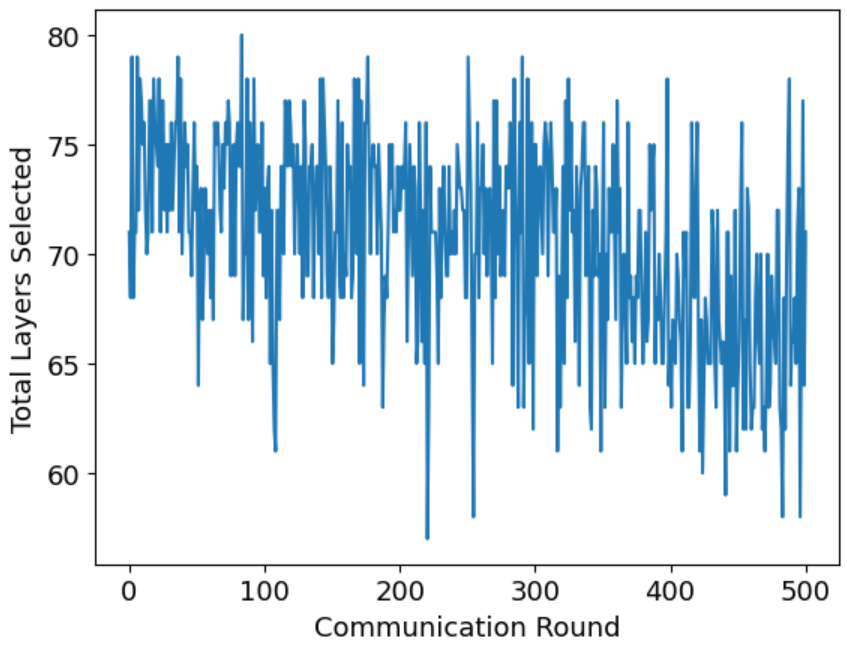} 
    \label{fd002_ls_30}
}
\hfill
\subfloat[][FD003]{
    \includegraphics[width=0.23\textwidth]{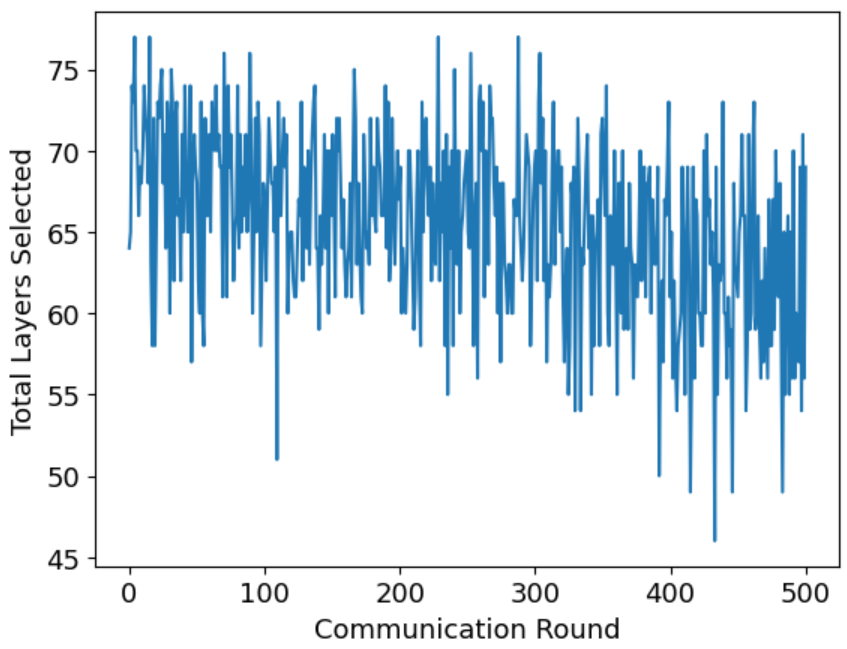} 
    \label{fd003_ls_30}
}
\hfill
\subfloat[][FD004]{
    \includegraphics[width=0.23\textwidth]{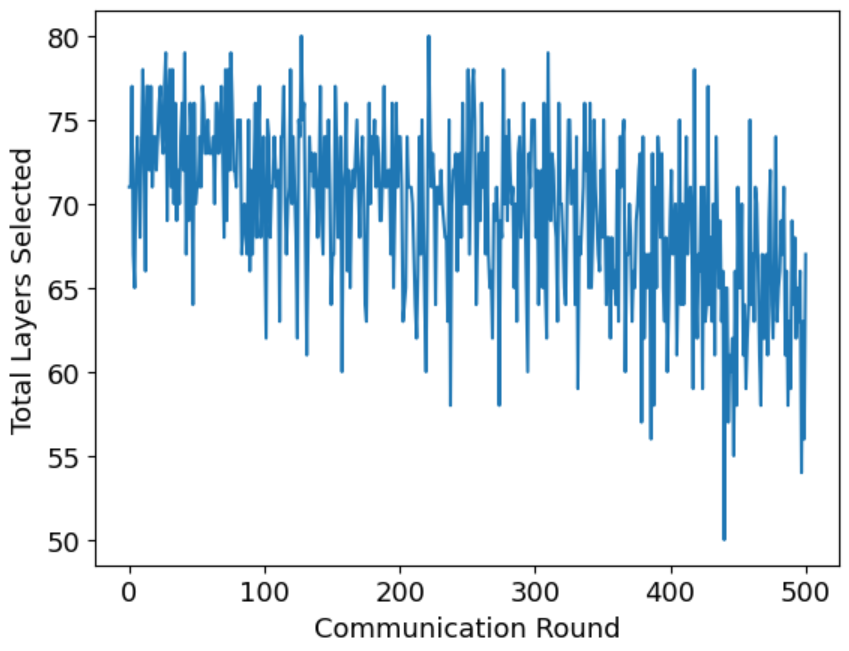}
    \label{fd004_ls_30}
}
\caption{\label{Plots_ls_across_datasets_30}
Comparison of layer selection actions chosen by RL-FL client agents across \textbf{FD001–FD004} at a fixed packet loss rate of \textbf{30\%}. Each subplot shows the evolution of $\sum_{n=1}^{20} ls_n^t$ over communication rounds for the corresponding dataset.}
\end{figure*}

Fig.~\ref{Plots_ls_across_datasets_30} compares layer selection actions across FD001--FD004 at 30\% packet loss rate. All datasets show stable trends with mild late-round decreases, though the degree of variability differs. FD002 is the most stable, FD001 and FD004 show more fluctuation, and FD003 exhibits somewhat lower average values. Overall, layer selection policies generalize more uniformly across datasets than retransmission policies.

\subsection{Analysis of Communication Cost}
\begin{figure*}[!t]
\centering
\subfloat[][FD001]{
    \includegraphics[width=0.23\textwidth]{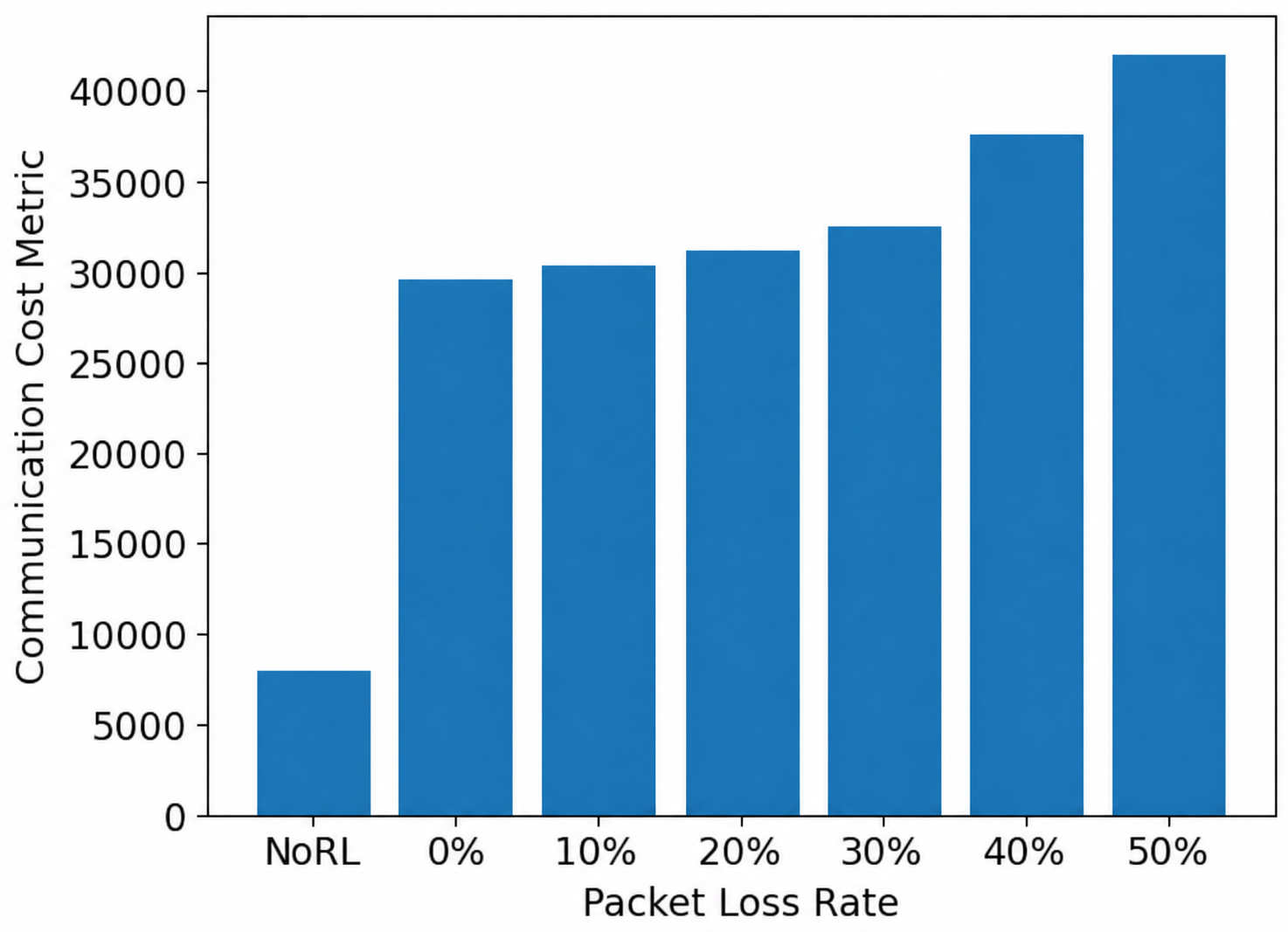}
    \label{fd001_commcost}
}
\hfill
\subfloat[][FD002]{
    \includegraphics[width=0.23\textwidth]{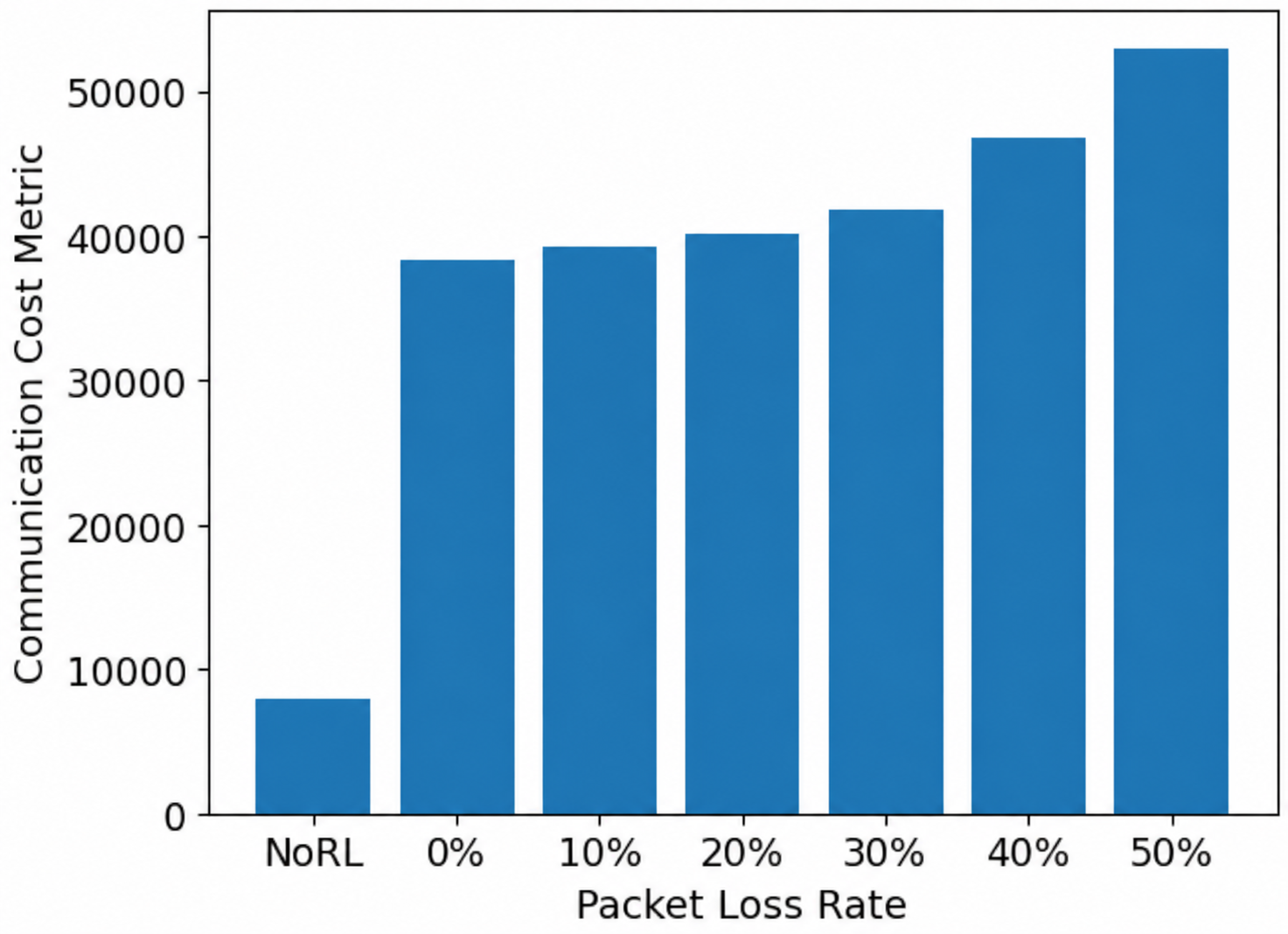}
    \label{fd002_commcost}
}
\hfill
\subfloat[][FD003]{
    \includegraphics[width=0.23\textwidth]{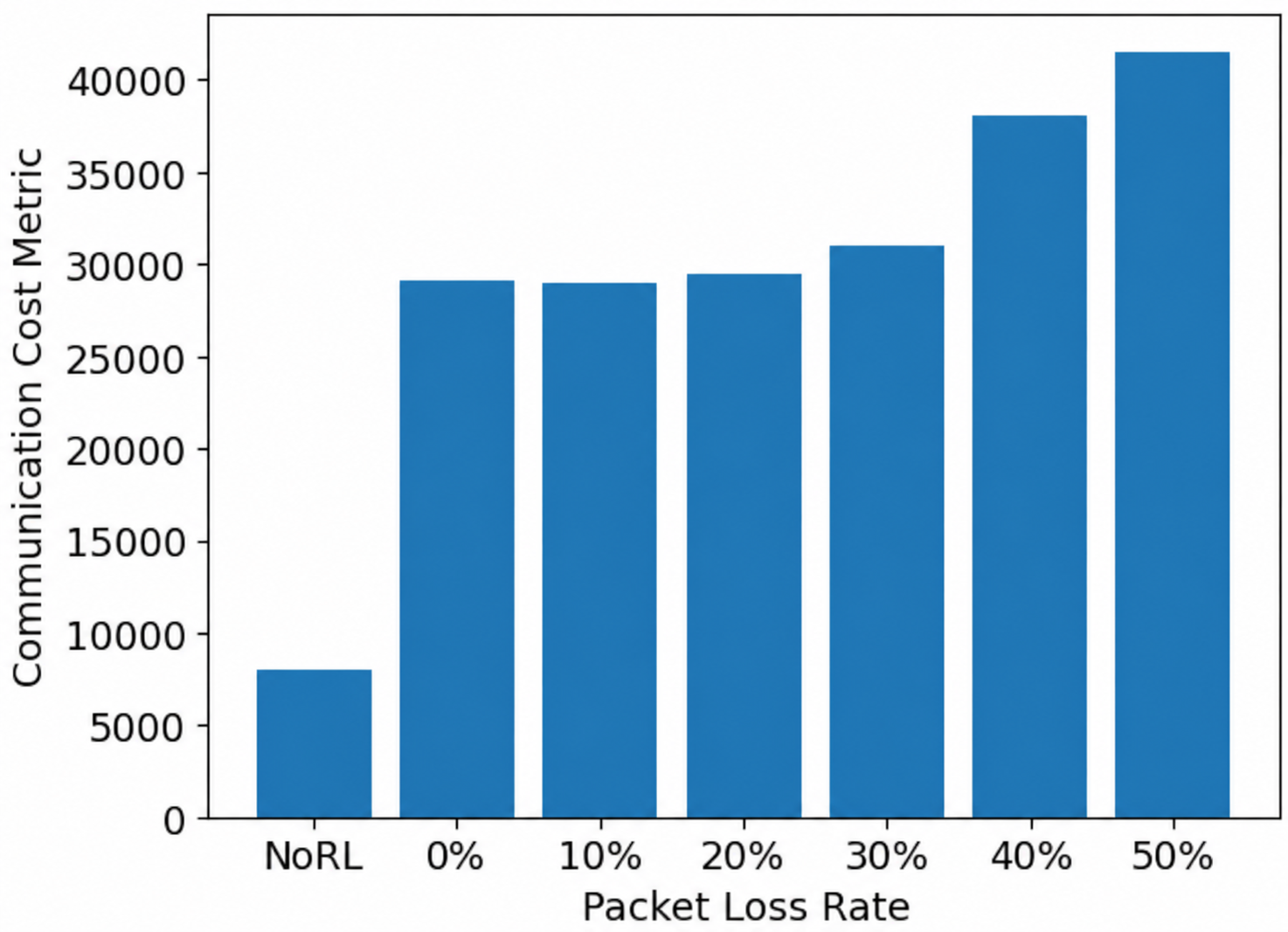}
    \label{fd003_commcost}
}
\hfill
\subfloat[][FD004]{
    \includegraphics[width=0.23\textwidth]{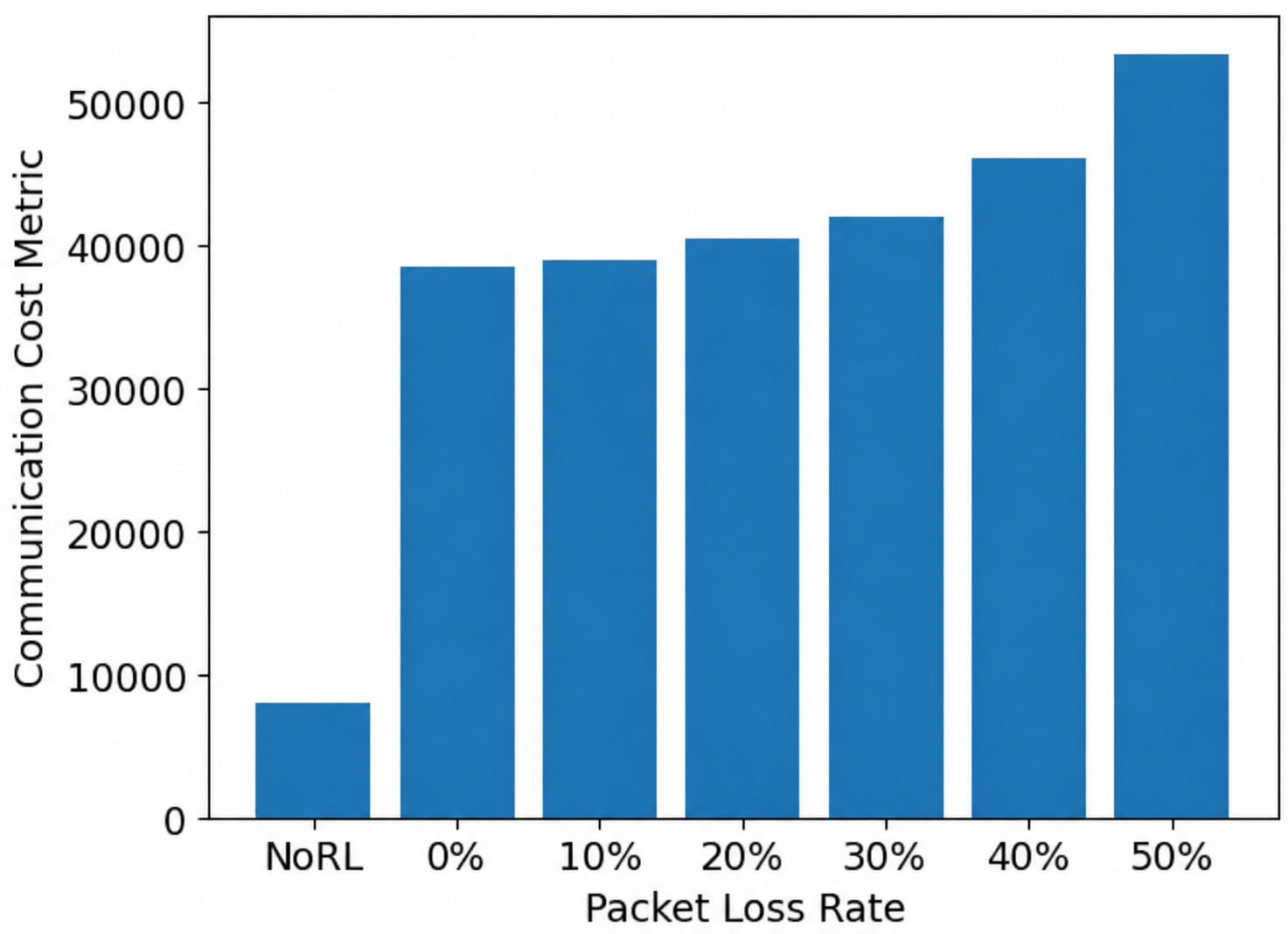}
    \label{fd004_commcost}
}
\caption{\label{Plots_commcost_across_datasets}
Total communication load vs.\ packet loss rate for \textbf{FD001–FD004}. Each subplot reports 
$\sum_{t=1}^{T}\sum_{n=1}^{20} ac(R_n^t)\,ls_n^t$, 
where $T$ is the number of communication rounds, $ac(R_n^t)$ is the actual number of retransmissions used by client $n$ at round $t$, and $ls_n^t$ is the number of layers selected.}

\end{figure*}

Fig.~\ref{Plots_commcost_across_datasets} shows the total communication load across FD001--FD004 as packet loss rate increases. All datasets exhibit monotonic growth in communication load with higher loss rates, reflecting the greater number of retransmissions triggered by the MARL policy. We have used NoRL to denote the communication cost of standard federated averaging. In standard federated averaging, every layer is selected ($ls^{t}_{n} = 4, \forall (t, n)$) and transmitted exactly once during communication ($ac(R^{t}_{n}) = 1, \forall (t, n)$). FD001 has the lowest overall load, while FD002 and FD004 carry the highest values, exceeding 50,000 at 50\% loss. FD003 shows the steepest relative increase, more than doubling from its baseline to the 50\% loss case. These differences indicate that communication overhead is not uniform across datasets. While MARL effectively adapts to channel degradation in all cases, the magnitude of the required transmissions depends on dataset-specific factors such as variability in sensor signals and model convergence dynamics.

\section{Ablation Study} \label{SecAblation}

To better understand the contribution of each action type, we perform an ablation study on FD001 and FD002 at 10\% and 30\% packet loss. We compare three variants of the MARL framework: 

\begin{itemize}
    \item \textbf{RL-FL-RTx}: Only retransmission control is used, with all layers always updated ($ls_n^t=4$).
    \item \textbf{RL-FL-LS}: Only layer selection is used, with retransmissions fixed at $R_n^t=1$. This variant also includes the action $ls_n^t=0$, meaning a client may skip participation in a round.
    \item \textbf{RL-FL}: The full method, where both retransmission control and layer selection are jointly optimized.
\end{itemize}

Table \ref{tab:ablation_fd001_fd002} presents the results. 

\begin{table}[h!]
\caption{Ablation study on FD001 and FD002: Performance (MSE) of RL-FL variants at 10\% and 30\% packet loss rate.}
\label{tab:ablation_fd001_fd002}
\centering
\begin{tabular}{|c|c||c|c|c|c|}
\hline
Dataset & Loss & FL & RL-FL & RL-FL-RTx & RL-FL-LS \\
\hline\hline
FD001 & $10\%$ & \textbf{18.07} & 18.58 & 18.27 & 18.65 \\
FD001 & $30\%$ & 20.52 & 19.85 & \textbf{18.51} & 19.97 \\
FD002 & $10\%$ & 32.61 & \textbf{30.31} & 30.44 & 30.97 \\
FD002 & $30\%$ & 34.35 & \textbf{32.06} & 30.52 & 31.19 \\
\hline
\end{tabular}
\end{table}

\begin{figure*}[!t]
\centering
\subfloat[][RL-FL.]{
    \includegraphics[width=0.31\textwidth]{RL_FL_0.3_500_20.png} 
    \label{fd001_30_rlfl}
}
\hfill
\subfloat[][RL-FL-RTx.]{
    \includegraphics[width=0.31\textwidth]{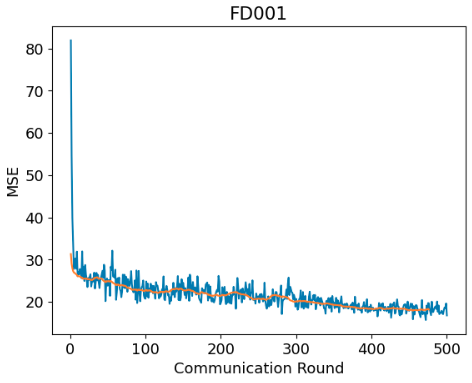}
    \label{fd001_30_rtx}
}
\hfill
\subfloat[][RL-FL-LS.]{
    \includegraphics[width=0.31\textwidth]{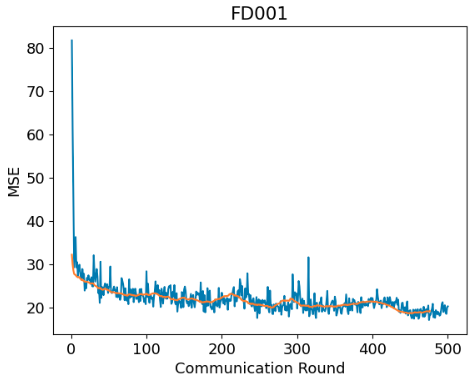}
    \label{fd001_30_ls}
}
\caption{\label{Plots_fd001_ablation_30}
Convergence plots of RL-FL variants on \textbf{FD001} for a $30\%$ packet loss rate. The plots compare the full RL-FL algorithm, RL-FL-RTx (retransmission only), and RL-FL-LS (layer selection only).}
\end{figure*}

From the ablation results in Table~\ref{tab:ablation_fd001_fd002} and the convergence curves in Fig.~\ref{Plots_fd001_ablation_30}, several trends become clear. At 10\% packet loss rate, the FL baseline achieves the lowest MSE on FD001, while full RL-FL performs best on FD002. At 30\% loss rate, RL-FL-RTx consistently achieves the lowest final MSE on both datasets, showing that retransmission control is the dominant factor under severe channel degradation. RL-FL-LS also improves over FL but is less effective, particularly at higher loss levels.

The convergence plots further highlight qualitative differences in learning dynamics. RL-FL produces the smoothest convergence curve with much less jitter, while RL-FL-RTx, despite reaching the lowest final error, exhibits more variance during training. RL-FL-LS shows the noisiest and weakest convergence behavior. These observations suggest that while retransmission control is most effective in reducing error at high packet loss rates, the full RL-FL achieves more stable and consistent training trajectories.

It is worth noting that although RL-FL also controls retransmission actions, its larger joint action space (combining retransmissions and layer selections) makes the learning problem more complex. This added difficulty can sometimes result in RL-FL underperforming RL-FL-RTx in terms of final MSE at high loss rates, even though RL-FL is more flexible in principle and offers smoother convergence.

\section{Conclusion and Future Work} \label{SecConc}

In this work, we have proposed a MARL based framework to mitigate the effect of wireless loss when training a RUL prediction model using FL over a lossy wireless network. The client RL agents were trained using the QMIX based value function factorization approach. Our simulations have shown the effectiveness of this approach as a significant drop in the final RUL prediction error was observed. All models were trained using the communication-cost-aware reward (equation \ref{groupReward}), which balances accuracy improvement with communication efficiency. Our experiments on the CMAPSS dataset demonstrate that RL-FL consistently improves performance under noisy channels. Gains are most pronounced on FD002 and FD003, where RL-FL achieves double-digit percentage improvements at high packet loss rates. FD001 also shows clear benefits at higher packet loss rates, while FD004 exhibits mixed behavior, suggesting dataset-specific sensitivity.

Overall, these results show that learning how much to transmit (layer subset selection) and how robustly to transmit (number of retransmissions) is an effective countermeasure to wireless loss in FL training.

In future work, we plan to further hierarchically factorize the value function w.r.t. different action types like layer selection and number of retransmissions, to reduce action space dimensions, which may improve convergence speed and further enhance RUL prediction performance. Furthermore, we also plan to train RL agents which are sensitive to variable packet loss rates in actual wireless networks. In addition, we will also conduct studies to compare the performance of different FL algorithms in combination with other state-of-the-art MARL algorithms in noisy communication channels.

\bibliographystyle{IEEEtran}
\bibliography{bibliography.bib}

@article{lee2014prognostics,
  title={Prognostics and health management design for rotary machinery systems - Reviews, methodology and applications},
  author={Lee, Jay and Wu, Fangji and Zhao, Wenyu and Ghaffari, Masoud and Liao, Linxia and Siegel, David},
  journal={Mechanical {S}ystems and {S}ignal {P}rocessing},
  volume={42},
  number={1-2},
  pages={314--334},
  year={2014},
  publisher={Elsevier}
}

@article{vaswani2017attention,
  title={Attention is all you need},
  author={Vaswani, Ashish and Shazeer, Noam and Parmar, Niki and Uszkoreit, Jakob and Jones, Llion and Gomez, Aidan N and Kaiser, {\L}ukasz and Polosukhin, Illia},
  journal={Advances in {N}eural {I}nformation {P}rocessing {S}ystems},
  volume={30},
  year={2017}
}

@article{devlin2018bert,
  title={B{E}{R}{T}: Pre-training of deep bidirectional transformers for language understanding},
  author={Devlin, Jacob and Chang, Ming-Wei and Lee, Kenton and Toutanova, Kristina},
  journal={arXiv preprint arXiv:1810.04805},
  year={2018}
}

@inproceedings{dong2018speech,
  title={Speech-transformer: a no-recurrence sequence-to-sequence model for speech recognition},
  author={Dong, Linhao and Xu, Shuang and Xu, Bo},
  booktitle={2018 IEEE International Conference on Acoustics, Speech and Signal Processing (ICASSP)},
  pages={5884--5888},
  year={2018},
  organization={IEEE}
}

@inproceedings{zeng2022deciwatch,
  title={Deciwatch: A simple baseline for 10$\times$ efficient 2d and 3d pose estimation},
  author={Zeng, Ailing and Ju, Xuan and Yang, Lei and Gao, Ruiyuan and Zhu, Xizhou and Dai, Bo and Xu, Qiang},
  booktitle={European Conference on Computer Vision},
  pages={607--624},
  year={2022},
  organization={Springer}
}

@article{wen2022transformers,
  title={Transformers in time series: A survey},
  author={Wen, Qingsong and Zhou, Tian and Zhang, Chaoli and Chen, Weiqi and Ma, Ziqing and Yan, Junchi and Sun, Liang},
  journal={arXiv preprint arXiv:2202.07125},
  year={2022}
}

@article{li2019enhancing,
  title={Enhancing the locality and breaking the memory bottleneck of transformer on time series forecasting},
  author={Li, Shiyang and Jin, Xiaoyong and Xuan, Yao and Zhou, Xiyou and Chen, Wenhu and Wang, Yu-Xiang and Yan, Xifeng},
  journal={Advances in {N}eural {I}nformation {P}rocessing {S}ystems},
  volume={32},
  year={2019}
}

@inproceedings{zhou2021informer,
  title={Informer: Beyond efficient transformer for long sequence time-series forecasting},
  author={Zhou, Haoyi and Zhang, Shanghang and Peng, Jieqi and Zhang, Shuai and Li, Jianxin and Xiong, Hui and Zhang, Wancai},
  booktitle={Proceedings of the AAAI Conference on Artificial Intelligence},
  volume={35},
  number={12},
  pages={11106--11115},
  year={2021}
}

@article{wu2021autoformer,
  title={Autoformer: Decomposition transformers with auto-correlation for long-term series forecasting},
  author={Wu, Haixu and Xu, Jiehui and Wang, Jianmin and Long, Mingsheng},
  journal={Advances in Neural Information Processing Systems},
  volume={34},
  pages={22419--22430},
  year={2021}
}

@inproceedings{liu2021pyraformer,
  title={Pyraformer: Low-complexity pyramidal attention for long-range time series modeling and forecasting},
  author={Liu, Shizhan and Yu, Hang and Liao, Cong and Li, Jianguo and Lin, Weiyao and Liu, Alex X and Dustdar, Schahram},
  booktitle={International Conference on Learning Representations},
  year={2021}
}

@article{cirstea2022triformer,
  title={Triformer: Triangular, Variable-Specific Attentions for Long Sequence Multivariate Time Series Forecasting--Full Version},
  author={Cirstea, Razvan-Gabriel and Guo, Chenjuan and Yang, Bin and Kieu, Tung and Dong, Xuanyi and Pan, Shirui},
  journal={arXiv preprint arXiv:2204.13767},
  year={2022}
}

@inproceedings{zhou2022fedformer,
  title={F{E}{D}former: Frequency enhanced decomposed transformer for long-term series forecasting},
  author={Zhou, Tian and Ma, Ziqing and Wen, Qingsong and Wang, Xue and Sun, Liang and Jin, Rong},
  booktitle={International Conference on Machine Learning},
  pages={27268--27286},
  year={2022},
  organization={PMLR}
}

@inproceedings{chattopadhyay2024position,
  title={A Position Aware Transformer Architecture for Traffic State Forecasting},
  author={Chattopadhyay, Rajarshi and Tham, Chen-Khong},
  booktitle={2024 IEEE 99th Vehicular Technology Conference (VTC2024-Spring)},
  year={2024},
  organization={IEEE}
}

@article{bruna2014spectral,
  title={Spectral networks and locally connected networks on graphs},
  author={Bruna, Joan and Zaremba, Wojciech and Szlam, Arthur and LeCun, Yann},
  journal={International Conference on Learning Representations (ICLR 2014)},
  year={2014}
}

@article{bronstein2017geometric,
  title={Geometric deep learning: going beyond {E}uclidean data},
  author={Bronstein, Michael M and Bruna, Joan and LeCun, Yann and Szlam, Arthur and Vandergheynst, Pierre},
  journal={IEEE Signal Processing Magazine},
  volume={34},
  number={4},
  pages={18--42},
  year={2017},
  publisher={IEEE}
}

@inproceedings{mcmahan2017communication,
  title={Communication-efficient learning of deep networks from decentralized data},
  author={McMahan, Brendan and Moore, Eider and Ramage, Daniel and Hampson, Seth and y Arcas, Blaise Aguera},
  booktitle={Artificial Intelligence and Statistics},
  pages={1273--1282},
  year={2017},
  organization={PMLR}
}

@article{sunehag2017value,
  title={Value-decomposition networks for cooperative multi-agent learning},
  author={Sunehag, Peter and Lever, Guy and Gruslys, Audrunas and Czarnecki, Wojciech Marian and Zambaldi, Vinicius and Jaderberg, Max and Lanctot, Marc and Sonnerat, Nicolas and Leibo, Joel Z and Tuyls, Karl and others},
  journal={arXiv preprint arXiv:1706.05296},
  year={2017}
}

@article{rashid2020monotonic,
  title={Monotonic value function factorisation for deep multi-agent reinforcement learning},
  author={Rashid, Tabish and Samvelyan, Mikayel and De Witt, Christian Schroeder and Farquhar, Gregory and Foerster, Jakob and Whiteson, Shimon},
  journal={Journal of Machine Learning Research},
  volume={21},
  number={178},
  pages={1--51},
  year={2020}
}

@inproceedings{saxena2008damage,
  title={Damage propagation modeling for aircraft engine run-to-failure simulation},
  author={Saxena, Abhinav and Goebel, Kai and Simon, Don and Eklund, Neil},
  booktitle={2008 International Conference on Prognostics and Health Management},
  pages={1--9},
  year={2008},
  organization={IEEE}
}

@INPROCEEDINGS{chattopadhyay2025graph,
  author={Chattopadhyay, Rajarshi and Tham, Chen-Khong},
  booktitle={2025 IEEE 101st Vehicular Technology Conference (VTC2025-Spring)}, 
  title={Marl-{E}nhanced {F}ederated {L}earning for {P}redicting {R}emaining {U}seful {L}ifetime}, 
  year={2025},
  volume={},
  number={},
  pages={1-6},
  doi={10.1109/VTC2025-Spring65109.2025.11174527}}

\end{document}